\documentclass[11pt]{article}

\usepackage{lipsum}
\usepackage[numbers]{natbib}
\usepackage{amsmath, amssymb, amsfonts, amsopn}
\usepackage{graphicx}
\usepackage{epstopdf}
\usepackage{algorithmic}
\usepackage{algorithm}
\usepackage{adjustbox}
\usepackage{hhline}
\usepackage{bm}
\usepackage{bbm}
\usepackage{mathtools}      
\usepackage{nicefrac}       
\usepackage{microtype}      
\usepackage{xcolor}         
\usepackage{booktabs}       
\usepackage{enumitem}       
\usepackage[margin=1in]{geometry}

\usepackage{tikz}
\usetikzlibrary{bayesnet}
\usetikzlibrary{arrows, calc, backgrounds, shapes.geometric, decorations.pathreplacing, angles, quotes}
\usepackage{etoolbox}  

\newcommand{\appsection}[1]{
  \refstepcounter{section}
  \section*{Appendix \Alph{section}: #1}
  \addcontentsline{toc}{section}{Appendix \Alph{section}: #1}
  \label{appendix\Alph{section}}
}

\usepackage[utf8]{inputenc}
\usepackage[T1]{fontenc}
\usepackage{url}
\usepackage{hyperref}
\usepackage{amsthm}
\theoremstyle{definition}
\newtheorem{definition}{Definition}
\newtheorem{remark}{Remark}
\newtheorem{proposition}{Proposition}
\newtheorem{theorem}{Theorem}
\ifpdf
  \DeclareGraphicsExtensions{.pdf,.png,.jpg,.eps}
\else
  \DeclareGraphicsExtensions{.eps}
\fi

\DeclarePairedDelimiterX{\infdivx}[2]{(}{)}{#1\;\delimsize\|\;#2}
\newcommand{\kldiv}{\textnormal{KL}\infdivx}

\newcommand{\bbr}{\mathbb{R}}
\newcommand{\bbn}{\mathbb{N}}
\newcommand{\bbe}{\mathbb{E}}

\newcommand{\inc}{\trianglelefteq}
\newcommand{\dash}{$-$}

\newcommand{\wt}{\widetilde}
\newcommand{\T}{\mathsf{T}}

\newcommand{\mcf}{\mathcal{F}}
\newcommand{\mcg}{\mathcal{G}}

\newcommand{\mch}{\mathcal{H}}
\newcommand{\mcl}{\mathcal{L}}
\newcommand{\cc}{\mathcal{C}}

\DeclareMathOperator{\diag}{diag}

\DeclareMathOperator{\vect}{vec}
\DeclareMathOperator{\diam}{diam}

\DeclareMathOperator{\tr}{tr}

\DeclareMathOperator{\MLP}{MLP}

\DeclareMathOperator{\Cauchy}{Cauchy}
\usepackage{lastpage}
\title{Bayesian Sheaf Neural Networks}

\author{
  Patrick Gillespie \\
  Department of Mathematics, University of Tennessee, Knoxville, TN, USA \\
  \texttt{pgilles5@vols.utk.edu}
  \and
  Layal Bou Hamdan \\
  Department of Mathematics, University of Tennessee, Knoxville, TN, USA \\
  \texttt{lbouhamd@vols.utk.edu}
  \and
  Ioannis Schizas \\
  DEVCOM ARL, Army Research Lab, Aberdeen, MD, USA\\ 
  \texttt{ioannis.d.schizas.civ@army.mil}
  \and
David L. Boothe\\
DEVCOM ARL, Army Research Lab, Aberdeen, MD, USA\\ 
  \texttt{ david.l.boothe7.civ@army.mil}
      \and
  Vasileios Maroulas\\
  Department of Mathematics, University of Tennessee, Knoxville, TN, USA\\  \texttt{vasileios.maroulas@utk.edu}
}

\date{} 
\begin{document}

\maketitle

\section*{Abstract}
    Equipping graph neural networks with a convolution operation defined in terms of a \emph{cellular sheaf} offers advantages for learning expressive representations of heterophilic graph data. The most flexible approach to constructing the sheaf is to learn it as part of the network as a function of the node features.
    However, this leaves the network potentially overly sensitive to the learned sheaf. As a counter-measure, we propose a variational approach to learning cellular sheaves within sheaf neural networks, yielding an architecture we refer to as a \emph{Bayesian sheaf neural network}. As part of this work, we define a novel family of reparameterizable probability distributions on the rotation group $SO(n)$ using the Cayley transform. We evaluate the Bayesian sheaf neural network on several graph datasets, and show that our Bayesian sheaf models
    achieve leading performance compared to baseline models and are less sensitive to the choice of hyperparameters under limited training data settings.

\vspace{1em}
\noindent\textbf{Keywords:} 
  Cellular sheaf, Bayesian neural network, variational inference, Cayley transform, uncertainty quantification.

\section{Introduction}
Geometric and topological deep learning are rapidly developing research areas which, broadly speaking, aim to use geometric priors about the data of interest to develop neural network architectures using methods and tools from differential geometry, topology, and graph theory \citep{Bronstein2021GeometricDL, papamarkou2024topological}. 
Graph neural networks are 
a central method in geometric deep learning and have been applied to a wide range of tasks such as recommender systems \citep{ying2018graph}, traffic modeling \citep{li2017diffusion, yu2017spatio}, and predicting properties of chemicals \citep{gilmer2017neural}.  
Building on the basic idea of graph neural networks, architectures such as simplicial 
and hypergraph neural networks extend the graph convolutions 
to 
higher-order
operations, which have advantages when the dataset contains salient relationships 
among groups of nodes rather than just node pairs, 
as in citation networks \citep{ebli2020simplicial,  feng2019hypergraph}, chemical data \citep{bodnar2021weisfeiler, neurips-bodnar2021b}, and neural data \citep{mitchell2024topological}. However, such convolution operators suffer from over-smoothing, and thus tend to perform best on homophilic graph data, where connected nodes 
have a high likelihood of belonging to the same class \citep{rusch2023survey, bodnar2022neural}. 

Several approaches have been taken to address the limited capabilities of GCNs and higher-order variants for heterophilic graph data (i.e., not homophilic) \citep{yan2021two, pei2020geom, zhu2020beyond, zhao2019pairnorm}. One approach replaces the graph 
convolution operation 
with one defined in terms of a \emph{cellular sheaf} \citep{hansen2020sheaf, bodnar2022neural}. 
A cellular sheaf on a graph assigns a vector space to each vertex and each edge, along with linear maps, known as \emph{restriction maps}, that relate the data on edges to the data on their incident vertices. 
Restriction maps provide additional flexibility in the message passing operation. They allow neighboring nodes to maintain distinct features throughout the layers of the sheaf neural network: an advantageous quality when working with heterophilic graph data.

While sheaf neural networks were developed to address the issue of over-smoothing in graph neural networks, they are still susceptible to many of the issues common to all neural networks, such as overfitting to limited training data and lack of robustness to noisy data, weight initializations, and hyperparameter selections. An approach to mitigating such issues is to apply tools and ideas from Bayesian statistics to deep learning \citep{jospin2022hands, papamarkou2024position}. 
Bayesian approaches to machine learning have been used to quantify the uncertainty in neural network predictions \citep{kendall2017uncertainties, gal2016dropout}, reduce overfitting \citep{gal2015bayesian, gal2016theoretically}, add robustness to adversarial attacks \citep{bortolussi2024robustness}, 
and to improve the overall predictive performance of machine learning algorithms \citep{fan2020bayesian, hajiramezanali2019variational, maroulas2022bayesian}.

In this paper, we apply a Bayesian framework to learning cellular sheaves within a sheaf neural network. Specifically, we consider the sheaf Laplacian as an object that enables the integration of the graph Laplacian with cellular sheaves, treating it as a latent random variable of the sheaf neural network.

\textbf{Our main contributions are as follows. }
\begin{enumerate}
    \item We introduce \textbf{Bayesian sheaf neural networks} (BSNN) in which the sheaf Laplacian operator of the sheaf neural network is a latent random variable whose distribution is parameterized by a function of the input data. Motivated by the evidence lower bound for variational Bayesian inference, we derive a pertinent loss function for our BSNN which contains a Kullback-Leibler (KL) divergence regularization term, 
    enabling efficient training. 
    \item We define a \textbf{novel family of reparameterizable probability distributions} on the rotation group $SO(n)$ that we refer to as \emph{Cayley distributions}. The Cayley distributions are reparameterizable and have tractable density functions, and thus can be used to perform variational inference on the sheaf Laplacian within the BSNN when the sheaf restriction maps belong to $SO(n)$. We derive closed-form expressions for the KL divergence of the Cayley distributions from the uniform distribution on $SO(n)$ when $n=2, 3$.
    \item We incorporate uncertainty quantification to rigorously evaluate our BSNN models on several benchmark graph datasets for node classification tasks, presenting extensive numerical results and comprehensive sensitivity analyses to validate robustness and performance.
\end{enumerate}
\subsection{Related Work}
We first discuss existing work on sheaf neural networks. Sheaf neural networks (SNN) were first introduced by \cite{hansen2020sheaf}, where they were shown to outperform graph convolutional networks on a semi-supervised node classification task using a hand-selected cellular sheaf. A more in-depth investigation of sheaf neural networks, sheaf diffusion processes, and the advantage of SNNs for heterophilic graph data was performed in \cite{bodnar2022neural}, where they also constructed a method of learning a suitable cellular sheaf from the data as part of the sheaf neural network. 

Building on these foundational approaches, several variants of SNNs have been proposed that extend their applicability or enhance their expressiveness.
They include SNNs with attention-based edge weights \citep{barbero2022attention} and SNNs defined for hypergraphs rather than cellular complexes \citep{duta2024sheaf}. In addition to hand-crafting a sheaf for a particular problem or learning a sheaf during training,  \cite{barbero2022connection} construct a sheaf at preprocessing time by assuming the graph data lie on a low-dimensional manifold and defining a cellular sheaf in a manner inspired by connections from differential geometry. 

While SNNs have not yet been explored through a Bayesian lens, uncertainty modeling and probabilistic reasoning have played an increasing role in geometric deep learning more broadly.
A variational auto-encoder using graph convolutional layers was considered in \cite{kipf2017graph} for an edge detection task.  \cite{davidson2018hyperspherical} discuss variational auto-encoders where the latent variable takes values in a hypersphere. \cite{hajiramezanali2019variational} combine graph variational autoencoders with graph recurrent networks to outperform existing state-of-the-art methods on dynamic link prediction tasks. Beyond variational autoencoders, \cite{fan2020bayesian} model attention weights in graph attention networks as random latent variables, as well as in other architectures which use attention. Given that the inclusion or exclusion of edges in graph datasets can be due to spurious noise or modeling assumptions, \cite{zhang2019bayesian} view graph datasets as realizations of parametric families of random graphs, and develop a Bayesian graph convolutional network accordingly.

A key component of a Bayesian formulation for sheaf neural networks is defining a distribution over the space of restriction maps, which often lie in the special orthogonal
$SO(n)$. The most well-known probability distributions on $SO(n)$ are the matrix Langevin (or von Mises-Fisher) distribution \citep{Pal2020Langevin, Chikuse2003Langevin} and the matrix Bingham distribution \citep{chikuse1990distributions}. However, such distributions are not easily reparameterizable, which makes incorporating them within neural networks trained via backpropagation difficult. 
While the matrix exponential used by \cite{falorsi2019reparameterizing} provides a reparameterizable mapping to $SO(d)$, it lacks tractable densities. 

In this paper, we define a family of reparameterizable distributions on $SO(n)$ using the Cayley transform which we refer to as Cayley distributions. These are distinct from the Cayley distributions defined in \cite{LEON2006412}, and we compare the two families of distributions in Section \ref{cayley_dist}. 
Moreover, \cite{Jauch_Hoff_Dunson_2020}
used a generalized version of the Cayley transform 
to simulate distributions on Stiefel and Grassmann manifolds, illustrating the broader applicability of Cayley-type methods in geometric statistics.

\subsection{Overview}
In Section \ref{prelims}, we review preliminary topics such as cellular sheaves and sheaf Laplacians. We review how the sheaf Laplacian can be used to define a sheaf neural network, and the connection to a sheaf diffusion process. We also review the Cayley distribution, variational inference, and several uncertainty quantification metrics. In Section \ref{bsnn}, we describe our variational approach to learning cellular sheaves in a sheaf neural network. We prove a result regarding the linear separation power of a sheaf diffusion process for sheaves consisting of special orthogonal maps. We define a reparameterizable family of distributions on $SO(n)$, which is used to define the variational distribution for cellular sheaves with special orthogonal restriction maps. In Section \ref{exp}, we evaluate our BSNN models on several node classification tasks and measure their predictive uncertainty. We end with a final discussion, as well as potential directions for future work in Section \ref{discussion}.
\section{Preliminaries}\label{prelims}
\subsection{Cellular Sheaves over Graphs}
\begin{definition}\label{Cellular Sheaf}
A \emph{cellular sheaf} $\mathcal{F}$ on an undirected graph $G = (V, E)$ assigns a vector space $\mathcal{F}(v)$ to each node $v \in V$, a vector space $\mathcal{F}(e)$ to each edge $e \in E$, and a linear map $\mathcal{F}_{v \inc e}: \mathcal{F}(v) \to \mathcal{F}(e)$ for each incident pair $v \inc e$ (i.e., node $v$ is incident to edge $e$).
\end{definition}
\begin{remark}
In the language of sheaf theory, $\mathcal{F}(v)$ is called the \emph{stalk} over $v$, and $\mathcal{F}_{v \inc e}$ is called a \emph{restriction map} \citep{bredon2012sheaf}.
\end{remark}
\begin{definition}\label{spaces}
The spaces of $0$-cochains and $1$-cochains are defined as the direct sums
\begin{equation*}
C^0(G; \mathcal{F}) := \bigoplus_{v \in V} \mathcal{F}(v) \quad \text{and} \quad C^1(G; \mathcal{F}) := \bigoplus_{e \in E} \mathcal{F}(e),
\end{equation*}
corresponding to the vector spaces assigned to the nodes and edges, respectively.
\end{definition}
\begin{definition}\label{Coboundary Map}
Let $x\in C^0(G;\mcf)$, $x_v$ denotes the value of $x$ at $\mcf(v)$. The coboundary map $\delta : C^0(G; \mathcal{F}) \to C^1(G; \mathcal{F})$ is defined componentwise by
\[
(\delta x)_e = \mathcal{F}_{v \inc e}(x_v) - \mathcal{F}_{u \inc e}(x_u),
\]
where $e$ is an edge oriented from $u$ to $v$ (this orientation is chosen arbitrarily to define a consistent sign convention; the graph itself remains undirected).
\end{definition}
\begin{definition}\label{Sheaf Laplacian}
The \emph{sheaf Laplacian} $L_{\mathcal{F}} : C^0(G; \mathcal{F}) \to C^0(G; \mathcal{F})$ is defined as
\begin{equation*}
L_{\mathcal{F}} := \delta^\top \delta.
\end{equation*}
\end{definition}
Given a sheaf $\mathcal{F}$ on a graph $G=(V,E)$ with $n$ nodes, and assuming that the stalk $\mathcal{F}(u)$ over each node $u \in V$ has the same dimension $d$, the sheaf Laplacian $L_{\mathcal{F}}$ can be represented as a block matrix of size $nd \times nd$.
The off-diagonal blocks $L_{ij} \in \mathbb{R}^{d \times d}$ are given by
$L_{ij} =-\mathcal{F}_{u_j \inc e}^\top \mathcal{F}_{u_i \inc e}  \text{ if there exists an edge } e \text{ between nodes } u_i \text{ and } u_j,$ and $0$ otherwise. The diagonal blocks are given by $L_{ii} = \sum_{u_i \inc e} \mathcal{F}_{u_i \inc e}^\top \mathcal{F}_{u_i \inc e},$ where the sum is taken over all edges $e$ incident to $u_i$.
\begin{remark}
Analogously to how message passing in graph convolutional networks uses the graph Laplacian 
$L = D - A$ \citep{kipf2017graph}, the sheaf Laplacian $L_{\mathcal{F}}$ defines message passing on cellular sheaves.
For further information on cellular sheaves and sheaf Laplacians, we refer the interested reader to \cite{HGh19}.
\end{remark}
\begin{definition}\label{Normalized Sheaf Laplacian}
The \emph{normalized $0$th sheaf Laplacian} $\Delta_{\mathcal{F}}$ associated with a cellular sheaf $\mathcal{F}$ over a graph $G$ is defined as
\begin{equation*}
\Delta_{\mathcal{F}} := D^{-1/2} L_{\mathcal{F}} D^{-1/2},
\end{equation*}
where $D$ is the block-diagonal matrix whose diagonal blocks are the $L_{ii}$ and whose off-diagonal blocks are zero matrices \citep{hansen2020sheaf}.
\end{definition}
\begin{remark}
The matrix $D$ is symmetric positive semi-definite, and therefore there exists a matrix $D^{1/2}$ such that $D^{1/2} D^{1/2} = D$. Moreover, we only consider 
restriction maps $\mathcal{F}_{u \inc e}$ that are invertible, 
ensuring $D$ is invertible and $D^{-1/2}$ is well-defined.
\end{remark}
\subsection{Sheaf Diffusion}\label{sheaf_dif}
Let $G$ be a graph with $n$ nodes and with $d${\dash}dimensional node features, and let $\mcf$ be a cellular sheaf over $G$ with $d${\dash}dimensional stalks. We represent the node features as an $nd${\dash}dimensional vector $X\in\bbr^{nd}$ so that the normalized sheaf Laplacian $\Delta_\mcf$, as an $nd\times nd$ block matrix, updates $X$ by matrix-vector multiplication $\Delta_\mcf X$. 
Motivated by how the graph convolutional layers of \cite{kipf2017graph} can be viewed as a discretization of a heat diffusion process $\tfrac{d}{dt}X(t)=-\Delta X(t)$ defined in terms of the normalized graph Laplacian $\Delta$, 
a similar diffusion process 
uses the sheaf Laplacian:
\begin{equation}\label{ode}
    X(0)=X, \quad \tfrac{d}{dt}X(t)=-\Delta_\mcf X(t).
\end{equation}
Applying the Euler method with unit time-steps to approximate solutions to Eq.(\ref{ode}) yields
\begin{equation}\label{time-step}
    X(t+1)=X(t)-\Delta_\mcf X(t).
\end{equation}
The authors of \cite{bodnar2022neural} use Eq.(\ref{time-step}) to define sheaf diffusion layers as follows. Allowing for $f\in\bbn$ distinct feature channels, let $X\in \bbr^{nd\times f}$ be a feature matrix. Given a sheaf Laplacian operator $\Delta_{\mcf(t)}$ for layer $t$, a sheaf diffusion layer is defined by setting
\begin{equation}\label{conv1}
X_{t+1}=X_t - \sigma\big(\Delta_{\mcf(t)}(I_n\otimes W_{t, 1})X_t W_{t, 2}\big).
\end{equation}
Here $W_{t, 1}\in \bbr^{d\times d}$ and $W_{t, 2}\in \bbr^{f\times f}$ are weight matrices for layer $t$, $\sigma$ is a nonlinear activation function, $\otimes$ denotes a Kronecker product. 

As a measure of the expressive power of the sheaf layers based on Eq.(\ref{time-step}), \cite{bodnar2022neural} study the ability of the sheaf diffusion process in Eq.(\ref{ode}) to linearly separate the features of nodes belonging to different classes.

\begin{definition}\label{power}
A class of sheaves $\mathcal{H}^d$ with $d${\dash}dimensional stalks is said to have \emph{linear separation power} over a class of labeled graphs $\mathcal{G}$ if for every $G\in \mathcal{G}$, there exists a sheaf $\mathcal{F}\in\mathcal{H}^d$ such that the sheaf diffusion process determined by $\mathcal{F}$ on $G$ linearly separates the features $X(t)$ of the classes of $G$ in the time limit $t\to\infty$, for almost all initial conditions $X(0)$.
\end{definition}

As part of a sheaf learning mechanism within a sheaf neural network, a fixed class of sheaves $\mch^d$ is parameterized, and the input data is mapped to a suitable sheaf $\mcf\in\mch^d$ through this parameterization.

For a node classification task with $C$ classes, the collection of sheaves $\mathcal{H}^d$ 
should ideally have linear separation power over the collection of connected graphs with $C$ classes. While one could always take $\mch^d$ to be as general as possible,
learning cellular sheaves whose restriction maps are diagonal or special orthogonal matrices involves fewer parameters. 
Recall that a special orthogonal matrix, that is, an element of $SO(n)$, is a matrix $Q$ such that $Q^T=Q^{-1}$ and $\det(Q)=1$.

The class of sheaves $\mathcal{H}^d_{\diag}$ with invertible diagonal restriction maps 
was shown to have linear separation power over the collection $\mathcal{G}_C$ of connected graphs with $C$ classes whenever $C\leq d$ \citep{bodnar2022neural}.  
However, for $\mathcal{H}_{so}^d$, the class of sheaves with $d\times d$ special orthogonal restriction maps, as discussed in Proposition 13 of \cite{bodnar2022neural}, only guarantees the existence of a $d$ such that $\mathcal{H}_{so}^d$ has linear separation power over $\mathcal{G}_C$ when $C\leq 8$.
\begin{remark}
 Although \cite{bodnar2022neural} consider sheaves with orthogonal restriction maps (i.e., elements of the orthogonal group $O(n)$), 
 the sheaf learning mechanism in their SNN is a continuous parametrization of
 $O(n)$ by the connected space $\bbr^m$, so it must have its image contained in a connected component of $O(n)$. 
 This implies that the restriction maps $\mcf_{u\inc e}$ will either all have determinant $1$ or all have determinant $-1$.
The maps $\mcf_{u\inc e}^\T\mcf_{v\inc e}$ used in the sheaf Laplacian must always have a determinant equal to $1$. Thus, it is effectively equivalent to learning restriction maps belonging to the special orthogonal group $SO(n)$. 
\end{remark}
\subsection{Cayley Transform}\label{cayley_prelim}
\begin{definition}
We denote the set of all $n\times n$ skew-symmetric matrices by $\mathfrak{so}(n)$. The \emph{Cayley transform}\citep{cayley} is a function $C:\mathfrak{so}(n)\to SO(n)$ defined by
\begin{equation}
    C(A)=(I-A)^{-1}(I+A)=2(I-A)^{-1}-I. 
\end{equation}
\end{definition}
If $A$ is an $n\times n$ skew-symmetric matrix, then $x^\T Ax=0$ for all $x\in\bbr^n$ due to the fact that $x^\T Ax = (x^\T Ax)^\T = -x^\T Ax$. Hence, if $(I-A)x=0$, then $x^\T(I-A)x=x^\T x=0$ implies that $x=0$. Thus, $I-A$ is invertible, showing that the Cayley transform is well-defined.
If $P\in SO(n)$ does not have $-1$ as an eigenvalue, $C^{-1}(P)$ exists and is given by 
\begin{equation}\label{cayley_eq}
    C^{-1}(P)=(P-I)(I+P)^{-1}=I-2(I+P)^{-1}.
\end{equation}
Hence $C$ is injective, and all but a measure zero subset of $SO(n)$ is contained in the image of $C$.
\begin{remark}
    The Cayley transform is sometimes alternatively defined as $C(A)=(I+A)^{-1}(I-A)=(I-A)(I+A)^{-1}$. Since the two definitions differ only by the precomposition of $C$ with the reflection $A\mapsto -A$, either choice of definition could be used without meaningful change.
\end{remark}

For a vector $\phi\in\bbr^{n(n-1)/2}$, let $X_\phi$ denote the skew symmetric matrix containing the entries of $\phi$ as its lower triangular entries. Explicitly, if $\phi=(\phi_1, \phi_2, \dots, \phi_{n(n-1)/2})$ and if $x_{ij}$ denotes the $i,j${\dash}th entry of $X_\phi$, then for $i>j$, set $x_{ij}=-x_{ji}=\phi_{ij-j(j+1)/2}$ and set $x_{ii}=0$. We may vectorize the Cayley transform and regard it as a map $\wt C:\bbr^{n(n-1)/2}\to\bbr^{n^2}$ by setting $\wt C(\phi)=\vect C(X_\phi)$. The Jacobian (determinant) of $\wt C$ is $J\wt C(\phi)=\det(D\wt C(\phi)^\T D\wt C(\phi))^{1/2}$, where $D\wt C(\phi)$ denotes the total derivative of $\wt C$ at the point $\phi$. Following \cite{LEON2006412}, the Jacobian $J\wt C(\phi)$ can be expressed in terms of $X_\phi$ by
\begin{equation}\label{jacobian}
    J\wt C(\phi) = \frac{2^{3n(n-1)/4}}{\det(I+X_\phi)^{n-1}}.
\end{equation}
\subsection{Variational Inference}\label{vi}

Let $P$ be a distribution on a measurable space $\mathcal{X}$ with density $p$ with respect to some reference measure $\mu$. If $X$ is a random variable with distribution $P$, we write $X \sim P$. To simplify notation, we will often refer to a distribution by its density function. 

For a function $f: \mathcal{X} \to \mathbb{R}$, we denote by $\mathbb{E}_P[f(X)]$ the expectation of $f(X)$ with respect to $P$, that is,
\begin{equation}
\mathbb{E}_P[f(X)] = \int_{\mathcal{X}} p(x) f(x) \, \mu(dx).
\end{equation}

The Kullback-Leibler (KL) divergence is a fundamental measure from information theory that quantifies how one probability distribution diverges from a second, reference distribution \citep{cover1999elements}. If $Q$ is another distribution over $\mathcal{X}$ with density $q$ (also with respect to $\mu$), the KL divergence of $P$ from $Q$ is defined as
\begin{equation}
\mathrm{KL}(P \,\|\, Q) = \int_{\mathcal{X}} p(x) \log \left( \frac{p(x)}{q(x)} \right) \mu(dx).
\end{equation}

Variational Inference (VI) transforms the problem of approximating intractable posterior distributions into an optimization task. Given a latent variable model with observations $y$ and latent variables $z$, the posterior $p(z|y)$ is often intractable due to the marginal likelihood requiring integration over the latent space. VI addresses this challenge by introducing a family of tractable distributions ${q_\phi(z)}$, parameterized by variational parameters $\phi$, and finding the member of this family that is closest to the true posterior in terms of the KL divergence:
\begin{equation}
\phi^*=\arg \min _\phi \operatorname{KL}\left(q_\phi(z) \| p(z \mid y)\right).
\end{equation}

This optimization is equivalent to maximizing the evidence lower bound (ELBO), which provides a lower bound on the marginal log-likelihood $\log p(y)$. The ELBO is given by
\begin{equation}
\mathcal{L}(\phi)=\mathbb{E}_{q_\phi(z)}[\log p(y, z)]-\mathbb{E}_{q_\phi(z)}\left[\log q_\phi(z)\right].
\end{equation}

Intuitively, the ELBO balances two competing objectives. The first term encourages $q_\phi(z)$ to assign high probability to latent variables $z$ that explain the observed data $y$ well. The second term acts as a regularizer, penalizing complexity in the approximate posterior by keeping it close to the prior. 

To see why the ELBO serves as a lower bound on $\log p(y)$, we note the identity:
\begin{equation}\label{elbo1}
\log p(y)=\mathcal{L}(\phi)+\mathrm{KL}\left(q_\phi(z) \| p(z \mid y)\right),
\end{equation}

which shows that maximizing the ELBO is equivalent to minimizing the divergence from the true posterior. Since the KL divergence is always nonnegative, it follows that $\log p(y) \geq \mcl(\phi)$.

In practice, the approximate posterior $q_\phi(z)$ is often chosen from a parametric family such as multivariate Gaussians with diagonal covariance, which allows reparameterization techniques and gradient-based optimization. For a comprehensive treatment of variational inference and its theoretical foundations, see \cite{blei2017variational}.

The choice of prior $p(z)$ plays an important role in variational inference. In many applications, a standard Gaussian prior $p(z) = \mathcal{N}(0, I)$ is used due to its simplicity and analytical tractability. Alternative priors, such as Laplace distributions for inducing sparsity or hierarchical priors for modeling structured uncertainty, can be used when specific inductive biases are desired. 
\subsection{Uncertainty Quantification Metrics}\label{uq_metrics}
The two main types of uncertainty in uncertainty quantification are \emph{aleatoric uncertainty} and \emph{epistemic uncertainty} \citep{derkiureghian2009aleatory}. Aleatoric uncertainty arises from inherent data or environmental randomness, such as sensor noise or measurement variability, and is therefore irreducible. In contrast, epistemic uncertainty comes from limited data or an incomplete model and is reducible through increased data or refined modeling. In machine learning, aleatoric uncertainty manifests as output ambiguity, whereas epistemic uncertainty reflects model confidence given data limitations \citep{huellermeier2021aleatoric}.

For a node classification task on a graph with $n$ nodes, $C$ classes, and an ensemble of $T$ stochastic forward passes, each input $x$ yields $T$ probability vectors  $p_t(y|x)$, where $ t = 1, \ldots, T$. The ensemble predictive distribution is obtained by averaging the predictions: $\bar{p}(y \mid x) = \frac{1}{T} \sum_{t=1}^{T} p_t(y \mid x)$. The predicted class $\hat{y}$ is the class with the highest mean probability, given by $\hat{y} = \arg\max_y \bar{p}(y \mid x)$, and the associated confidence is $\hat{p} = \max_y \bar{p}(y \mid x)$. The variance across ensemble predictions for each class $c \in \{1, \ldots, C\}$ captures the model's epistemic uncertainty, and is given by $\text{Var}_c(x) = \text{Var}_{t=1}^{T}\left[ p_t(y = c \mid x) \right].$ Several uncertainty qualification metrics can be defined as follows.

Predictive entropy measures the total uncertainty in a model's output (both aleatoric and epistemic). Lower entropy indicates that the model is more confident in its predictions \citep{gal2016dropout}. 
\begin{equation}\label{entropy}
H[\bar{p}(y|x)] = -\sum_{c=1}^{C} \bar{p}(y = c \mid x) \log \bar{p}(y = c \mid x).
\end{equation}

Epistemic variance measures the variability in model predictions due to uncertainty in model parameters. Lower epistemic variance implies that the model's predictions are consistent across stochastic forward passes \citep{lakshminarayanan2017simple}.
 \begin{equation}\label{ep_var}
\text{EpistemicVariance}(x) = \frac{1}{C} \sum_{c=1}^{C} \text{Var}_c(x).
\end{equation}

Mutual information (MI) quantifies the epistemic component of uncertainty by measuring the divergence between the entropy of the mean prediction and the mean of the individual entropies. Lower MI values indicate reduced epistemic uncertainty, suggesting strong agreement among ensemble members and greater confidence in the model’s parameters \citep{kendall2017uncertainties}.
 \begin{equation}\label{mi}
\text{MI}(x) = H\left[ \bar{p}(y|x) \right] - \frac{1}{T} \sum_{t=1}^{T} H\left[ p_t(y|x) \right].
\end{equation}

Expected Calibration Error (ECE) measures the discrepancy between predicted confidence scores and actual accuracy across a range of confidence intervals. To compute the ECE, the prediction confidences $\hat{p}_i$ are partitioned into $M$  equally spaced confidence intervals $I_m = \left( \frac{m-1}{M}, \frac{m}{M} \right]$ for $m = 1, \dots, M$. Let $B_m = \left\{ i \in \{1, \dots, N\} \,\middle|\, \hat{p}_i \in I_m \right\}$ be the set of indices of samples whose confidence scores fall into interval \(I_m\).

The \textit{accuracy} of bin \(B_m\) is defined as the proportion of correctly classified samples in that bin: $\text{acc}(B_m) = \frac{1}{|B_m|} \sum_{i \in B_m} \mathbbm{1}(\hat{y}_i = y_i),$ and the \textit{average confidence} is  $\text{conf}(B_m) = \frac{1}{|B_m|} \sum_{i \in B_m} \hat{p}_i.$

The Expected Calibration Error is computed as:
\begin{equation}\label{ece}
\text{ECE} = \sum_{m=1}^{M} \frac{|B_m|}{N} \left| \text{acc}(B_m) - \text{conf}(B_m) \right|.
\end{equation}

A lower ECE value indicates better calibration, meaning the predicted probabilities are more aligned with the observed accuracy \citep{guo2017calibration}.

\section{Bayesian Sheaf Neural Networks}\label{bsnn}
In this section, we introduce a Bayesian framework for Sheaf Neural Networks, extending the deterministic architecture by incorporating uncertainty in the learned sheaf structure.

\subsection{Problem Statement}
Consider a semi-supervised node classification problem, in which we have a graph dataset $G=(V, E)$ together with a node feature matrix $X$, a collection of observed node labels, $y$, and a collection of unobserved node labels, $y^*$, all of which are random samples.
We model the conditional probability $p_\theta(y | X)$ using a sheaf neural network whose layers are given by Eq.(\ref{conv1}), with network parameters $\theta$, adjacency matrix $A$, and a cellular sheaf $\mcf$.

Rather than taking the learned cellular sheaf $\mcf$
to be a deterministic function of $X$
, we regard $\mcf$ as a latent random variable of our model. See Figure \ref{fig:pgm} for a depiction of the probability graphical model. Note that the posterior distribution $p_\theta(\mcf|X, y)$ will be intractable due to the fact that the labels $y$ are modeled as the output of a neural network. 

To approximate posterior inference of $\mcf$ given the feature matrix $X$, the observed node labels $y$, and model parameters $\theta$, we assume that the intractable posterior distribution $p_\theta(\mcf|X, y)$ belongs to a predetermined family of variational distributions $q_\phi(\mcf|X, y)$ parameterized by weights $\phi$ as discussed in Section \ref{vi}. In our framework, we choose priors for $\mcf$ that reflect plausible variability in sheaf structures while preserving the tractability of inference.
\begin{figure}[H]\label{pgm-h}
\centering
  \tikz{
 \node[obs] (x) {$X$};
 \node[const, yshift=-1.75cm, inner sep=3pt] (A) {$A$};
 \node[const, yshift=-1cm, inner sep=3pt] (p) {$\phi$};
 \node[const, yshift=0.75cm, xshift=2cm, inner sep=3pt] (t) {$\theta$};
 \node[latent, yshift=-1.25cm, xshift=2cm] (F) {$\mcf$};
 \node[latent, xshift=4cm,path picture={\fill[gray!25] (path picture bounding box.south) rectangle (path picture bounding box.north west);}] (y) {$y$}; %
 \plate [inner sep=.25cm,yshift=.2cm] {plate1} {(F)} {$L$}; %
 \edge[style={-latex}]{x, F, t}{y} 
 \edge[style={-latex}]{x, A, p}{F}  
 }
 \caption{Probability graphical model of the Bayesian sheaf neural network for a semi-supervised node classification problem. The adjacency matrix of the graph is $A$, and $\theta$, $\phi$ are weights of the network.}
 \label{fig:pgm}
\end{figure}
\vspace{-1em}

Following Eq.(\ref{elbo1}), the log-likelihood $\log p_\theta(X, y)$ can be written
\begin{equation}\label{log-evidence}
\log p_\theta(X, y)=\kldiv{q_\phi(\mcf|X,y)}{p_\theta(\mcf|X,y)}+\mcl(\theta, \phi; X, y),
\end{equation}
where  
\begin{equation}\label{elbo-def}
\mcl(\theta, \phi; X, y):=\bbe_{q_\phi(\mcf|X,y)}\big[-\log q_\phi(\mcf|X,y) + \log p_\theta(\mcf, X, y)\big].
\end{equation}

In order to maximize $\mcl(\theta, \phi; X, y)$ with respect to the parameters $\theta, \phi$, we use the following equivalent expression:
\begin{equation}\label{elbo-eq}
\mcl(\theta, \phi; X, y)=\bbe_{q_\phi(\mcf|X, y)}\big[\log p_\theta(X, y|\mcf)\big]-\kldiv{q_\phi(\mcf|X,y)}{p_\theta(\mcf)}.
\end{equation}

In many cases, a closed-form expression can be derived for the KL divergence term in the right hand side of Eq.(\ref{elbo-eq}), leaving only the gradient of the expectation term with respect to $\theta$ and $\phi$ to be estimated. See Appendix \ref{app-kl} for calculations of the KL divergence in certain cases. 

Similar to \cite{kingma2013auto}, we estimate $\nabla_{\phi}\bbe_{q_\phi(\mcf|X, y)}[\log p_\theta(X, y|\mcf)]$ by making use of the so-called \emph{reparameterization trick}. 
In our case, for a random cellular sheaf $\mcf$ distributed as $\mcf \sim q_\phi(\mcf|X, y)$, the reparameterization trick expresses $\mcf$ as a deterministic function $\mcf=g_\phi(\epsilon, X)$ of random noise $\epsilon\sim p(\epsilon)$. This allows us to rewrite the expectation with respect to $q_\phi(\mcf|X, y)$ as an expectation with respect to $p(\epsilon)$ which in turn allows the gradient to be moved within the expectation. That is,
\begin{align*}
\nabla_{\phi}\bbe_{q_\phi(\mcf|X, y)}[\log p_\theta(X, y|\mcf)] &= \nabla_{\phi}\bbe_{p(\epsilon)}[\log p_\theta(X, y|g_\phi(\epsilon, X))]\\
&= \bbe_{p(\epsilon)}[\nabla_{\phi}\log p_\theta(X, y|g_\phi(\epsilon, X))]\\
&\approx \frac{1}{K}\sum_{k=1}^K \nabla_{\phi}\log p_\theta(X, y|g_\phi(\epsilon^{(k)}, X)),
\end{align*}
where $\epsilon^{(k)}$ are independent and identically distributed samples of $p(\epsilon)$ for $k\in \{1, 2, \dots, K\}$.

Hence, when the KL divergence term of Eq.(\ref{elbo-eq}) can be computed analytically, if $\mcf^{(k)}=g_\phi(\epsilon^{(k)}, X)$ for $\epsilon^{(k)}\sim p(\epsilon)$, our estimator is
\begin{equation}\label{loss_fn}
 \widetilde{\mcl}_\lambda(\theta, \phi; X, y) = \frac{1}{K}\sum_{k=1}^K \log p_\theta(X, y|\mcf^{(k)}) - \lambda \, \kldiv{q_\phi(\mcf|X,y)}{p_\theta(\mcf)}.
\end{equation}
Here, the weight $\lambda$ of the KL divergence term is included to combat issues with the KL divergence term vanishing during training, which we cyclically anneal during training in a manner similar to \cite{fu-etal-2019-cyclical}.
When the KL divergence term of Eq.(\ref{elbo-eq}) cannot be computed analytically, we may estimate it in a similar manner by noting that $\kldiv{q_\phi(\mcf|X,y)}{p_\theta(\mcf)}=\bbe_{q_\phi(\mcf|X,y)}[\log  (q_\phi(\mcf|X,y)/p_\theta(\mcf))]$. In this case, the estimator, also denoted by $\widetilde{\mcl}_\lambda$, is given by
\begin{equation}\label{loss_fn_est}
\widetilde{\mcl}_\lambda(\theta, \phi; X, y) = \frac{1}{K}\sum_{k=1}^K \Big(\log p_\theta(X, y|\mcf^{(k)})
- \lambda\log \frac{q_\phi(\mcf^{(k)}|X,y)}{p_\theta(\mcf^{(k)})}\Big),
\end{equation}
where again, $\mcf^{(k)}=g_\phi(\epsilon^{(k)}, X)$ for $\epsilon^{(k)}\sim p(\epsilon)$.

\subsection{Variational Sheaf Learner}\label{var_sheaf}
We make the simplifying assumption that the variational distribution $q_\phi(\mcf|X,y)$ from the previous section factors as the product $\prod_{u\inc e}q_\phi(\mcf_{u\inc e}|X, y)$ over all incident node-edge pairs $u\inc e$. While this assumption somewhat restricts the family of variational distributions, it makes computing the KL divergence term in the ELBO tractable. For further details, see Appendix \ref{app-kl}. 

We take each factor $q_\phi(\mcf_{u\inc e}|X, y)$ to be a distribution $Q(\mu_{u\inc e},\sigma_{u\inc e})$ which depends on the type of restriction map used within the sheaf, where the distributional parameters $\mu_{u\inc e},\sigma_{u\inc e}$ are a function of the node features $X$ and $\phi$ as follows. For each incident node-edge pair $u\inc e$, if $u'\inc e$ denotes the other node incident with $e$, a multi-layer perceptron with learned weights $\phi$, denoted $\MLP_\phi$, maps the concatenated vector $[x_u||x_{u'}]$ to distributional parameters $\mu_{u\inc e}$ and $\sigma_{u\inc e}$:
\begin{equation}\label{learner}
    [\,\mu_{u\inc e}\,||\,\sigma_{u\inc e}\,]=\MLP_\phi([x_u||x_{u'}]).
\end{equation}
Recall that a multi-layer perceptron is the composition of several feed-forward layers, each of which is a function of the form $x\mapsto \sigma(Wx+b)$, where $\sigma$ is a nonlinear activation function, in our case the exponential linear unit (ELU) \citep{elu}, $W$ is a weight matrix, and $b$ is a vector of biases. 

The distribution $Q$ for the variational sheaf learners is considered in three settings, where the learned sheaf $\mcf$ has restriction maps assumed to be: (1) general linear, (2) invertible diagonal, or (3) special orthogonal.

\vspace{0.5em}
\textbf{Case 1: General Linear.} In the case where the sheaf restriction maps $\mcf_{u\inc e}:\mcf(u)\to \mcf(e)$ are general linear maps, $\mu_{u\inc e}$ and $\sigma_{u\inc e}$ are $d^2${\dash}dimensional vectors. 
We take $Q(\mu_{u\inc e},\sigma_{u\inc e})$ to be normal with diagonal covariance so that for $z\sim \mathcal{N}_{d^2}(\mu_{u\inc e},\diag(\sigma_{u\inc e}))$ we set $\mcf_{u\inc e}$ to be the vector $z$ rearranged into a $d\times d$ matrix. The reparameterization trick in this case writes $z=\mu_{u\inc e} + \diag(\sigma_{u\inc e})\epsilon$ for $\epsilon\sim \mathcal{N}(\mathbf{0}, I_{d^2})$ so that $\mcf_{u\inc e}$, being the matrix obtained by reshaping the vector $z$, is a deterministic function of $\mu_{u\inc e}$, $\sigma_{u\inc e}$, and $\epsilon$.
We use a standard normal prior, where $p_\theta(\mcf)=\prod_{u\inc e}p_\theta(\mcf_{u\inc e})$ and $p_\theta(\mcf_{u\inc e})=\mathcal{N}_{d^2}(\mathbf{0}, I_{d^2})$.

\vspace{0.5em}
\textbf{Case 2: Invertible Diagonal.} In the case where sheaf restriction maps $\mcf_{u\inc e}$ are assumed to be invertible diagonal matrices with respect to the standard basis, $\mu_{u\inc e}$ and $\sigma_{u\inc e}$ are $d${\dash}dimensional vectors and we take $q_\phi(\mcf_{u\inc e}|X, y)=Q(\mu_{u\inc e},\sigma_{u\inc e})$ to be a normal distribution with diagonal covariance, 
so that $\mcf_{u\inc e}=\diag(z)$ for $z\sim \mathcal{N}_d(\mu_{u\inc e},\diag(\sigma_{u\inc e}))$. The reparameterization $\mcf=g_\phi(X,\epsilon)$ in this case is given by setting $\mcf_{u\inc e}=\diag(\mu_{u\inc e}+\sigma_{u\inc e}\epsilon)$ for $\epsilon\sim \mathcal{N}(\mathbf{0}, I_d)$.
Lastly, we use a standard normal prior, where $p_\theta(\mcf)=\prod_{u\inc e}p_\theta(\mcf_{u\inc e})$ and $p_\theta(\mcf_{u\inc e})=\mathcal{N}_d(\mathbf{0}, I_d)$. 

\vspace{0.5em}
\textbf{Case 3: Special Orthogonal.} 
The class of sheaves with special orthogonal restriction maps is of particular interest for sheaf neural networks, including our Bayesian sheaf neural networks, due to both theoretical and practical properties. From a theoretical point of view, certain results concerning the kernel of $\Delta_\mcf$ can be proven for sheaves using rotations as restriction maps \citep[Section 3.1]{bodnar2022neural}. Regarding the more practical aspects, rotations offer a nice middle-ground between diagonal and general linear maps in terms of balancing the number of parameters with the expressivity of the resulting sheaf. Moreover, when using rotations, the blocks along the diagonal of the sheaf Laplacian $L_\mcf$ are scalar multiples of the identity matrix. This makes computing the normalized sheaf Laplacian $\Delta_\mcf$ easier.

These properties motivate our focus on $SO(d)$ sheaves. To further support their utility, we extend previous theoretical results on their separation power as defined in Section \ref{sheaf_dif}.
In particular, we prove that for any number of classes $C\in\bbn$, we may find $d$ large enough such that $\mathcal{H}_{so}^d$ has linear separation power over $\mathcal{G}_C$. In the following proposition, $\lfloor\cdot\rfloor$ denotes the floor function.
\begin{proposition}\label{sheaf-power}
For any $d\geq 1$, if $\mathcal{G}$ denotes the class of connected graphs with $C\leq 7 \lfloor\tfrac{d}{4}\rfloor$ classes, then $\mathcal{H}_{so}^d$ has linear separation power over $\mathcal{G}$.
\end{proposition}
For a proof see Appendix \ref{app-sheaf-power}.
\begin{remark}
    An extra consequence of Proposition \ref{sheaf-power} is that it also provides a new upper bound on the stalk dimension $d$ necessary for $\mch^d_{gen}$, the collection of sheaves with $d\times d$ general linear restriction maps, to have linear separation power over the collection of connected graph $\mcg_C$ with $C$ classes. This is due to the simple fact that $\mch^d_{so}\subseteq \mch^d_{gen}$.
\end{remark}
We now turn to the practical challenge of incorporating these $SO(d)$ maps into the variational model. Typical probability distributions on $SO(d)$, such as the matrix Langevin distribution, are not easily reparameterizable. This makes incorporating such distributions into neural networks that are trained via back-propagation difficult. On the other hand, while reparameterizable distributions on $SO(d)$ have been constructed in \cite{falorsi2019reparameterizing} using the exponential map, the density functions of such distributions can be difficult to compute exactly, which obstructs one from computing, or even efficiently estimating, the KL divergence term in Eq.(\ref{elbo-eq}). 

In Section \ref{cayley_dist}, we introduce Cayley distributions,
a family of reparameterizable distributions $\cc_d(M, \kappa)$ on $SO(d)$ with closed form expressions for their probability density functions. 
We use $\cc_d(M,\kappa)$ as the variational distributions for sheaves with special orthogonal restriction maps. Hence when $\mcf_{u\inc e}\in SO(d)$, the distributional parameters are a mean rotation $\mu_{u\inc e}\in SO(d)$ and scalar concentration parameter $\sigma_{u\inc e}\in [0, 1)$, and we have $Q(\mu_{u\inc e},\sigma_{u\inc e})=\cc_d(\mu_{u\inc e},\sigma_{u\inc e})$. The reparameterization of the Cayley distributions is described in Remark \ref{cayley-rep}.
We use a prior distribution $p_\theta(\mcf)=\prod_{u\inc e}p_\theta(\mcf_{u\inc e})$ where each $p_\theta(\mcf_{u\inc e})$ is a uniform distribution on $SO(d)$.

\subsection[Cayley Distributions on $SO(n)$]{Cayley Distributions on $SO(n)$}\label{cayley_dist}
To enable reparameterizable inference on $SO(n)$, we define a novel family of distributions using the Cayley transform.
\begin{definition}
    Let $C:\mathfrak{so}(n)\to SO(n)$ denote the Cayley transform and let $X$ be a uniformly distributed random variable on $SO(n)$. Given parameters $M\in SO(n)$ and $0\leq \kappa<1$, we define the \emph{Cayley distribution} $\cc_n(M,\kappa)$ on $SO(n)$ inspired by the constructions in \cite{LEON2006412} to be the distribution of $Y=C\left(\frac{1-\kappa}{1+\kappa} C^{-1}(X)\right)M$.
\end{definition}
\begin{remark}\label{cayley-rep}
    Note that by the definition of the Cayley distribution, a random variable $Y\sim \cc_n(M,\kappa)$ can be expressed as a deterministic function of $M$, $\kappa$, and $X\sim U_{SO(n)}$. Hence these distributions are theoretically reparameterizable. For details on sampling from the uniform distribution on $SO(n)$, see for example \cite{orth_sampling}.
\end{remark}
In order to compute or efficiently estimate the KL divergence term in Eq.(\ref{loss_fn}), an expression for the density of $\cc_n(M,\kappa)$ is necessary. Using Eq.(\ref{jacobian}) for the Jacobian of the Cayley transform, we compute the density of $\cc_n(M,\kappa)$ in the following theorem.
\begin{theorem}\label{cayley-thm}
    Let $P \in SO(n)$, the Cayley distribution $\cc_n(M,\kappa)$ has density
    \begin{equation}\label{eq:Cal_dist}
            f_n(P;M, \kappa)=(1-\kappa^2)^{\frac{n(n-1)}{2}}\det\Big(PM^\T -\kappa I\Big)^{1-n},
    \end{equation}
    with respect to the normalized Haar measure on $SO(n)$.
\end{theorem}

For a proof see Appendix \ref{app-cayley}.

\begin{remark}
    The family of Cayley distributions defined in \cite{LEON2006412} have density proportional to $\det(PM^\T+I)^\kappa$ of Eq.\eqref{eq:Cal_dist}. Note that the concentration parameter $\kappa$ appears in the exponent of the determinant expression, whereas in our proposed family of Cayley distributions, $\kappa$ instead scales the identity matrix term within the determinant. This structural difference plays a key role in enabling reparameterizable constructions based on the Cayley transform, particularly in low dimensions.
\end{remark}
For small values of $n,$ the structure of $SO(n)$ allows for simpler parameterizations and more efficient sampling schemes. We now examine these cases to motivate practical approaches for facilitating differentiable sampling from $\cc_n(M,\kappa)$.

\textbf{Case $\mathbf{n=2}$. } If $P, M\in SO(2)$  are rotations by $\theta, \mu$ radians respectively, then $\cc_2(M,\kappa)$ has density
\begin{equation*}
    f_2(P;M, \kappa)=\frac{1}{2\pi}\cdot\frac{1-\kappa^2}{1+\kappa^2-2\kappa\cos(\theta-\mu)}.
\end{equation*}
This is the density of a wrapped Cauchy distribution on $SO(2)\cong S^1$, see \cite{kato2013extended}. Hence for the case $n=2$, the Cayley distributions coincide with known distributions. 

\textbf{Case $\mathbf{n=3}$. } Note that $SO(3)$ is homeomorphic to $3${\dash}dimensional real projective space $RP^3$, which is the quotient of $S^3$ under the identification of antipodal points. We show that $\cc_3(M,\kappa)$ is related to an \emph{angular central Gaussian} distribution \citep{tyler1987statistical} on $S^3$. Specifically, we show that $\cc_3(M,\kappa)$ is the pushforward of an angular central Gaussian distribution on $S^3$ under the quotient map $\Phi:S^3\to SO(3)$. This provides a reparameterizable sampling strategy, distinct from sampling via uniform $SO(3)$.

If $x\in\bbr^{n+1}$ is a unit vector, $\Lambda\in\bbr^{(n+1)\times (n+1)}$ is a symmetric positive definite matrix, and $\alpha_n$ denotes the volume of the $n${\dash}dimensional sphere $S^n$, the angular central Gaussian distribution $ACG_n(\Lambda)$ on $S^n$ has density
\begin{equation*}
    f(x; \Lambda)=\alpha_n^{-1}\det(\Lambda)^{-\frac{1}{2}}(x^\T \Lambda^{-1}x)^{-2},
\end{equation*}
with respect to the uniform measure on $S^n$.
\begin{proposition}\label{acg-prop}
    Let $\Phi:S^3\to \bbr\mathbb{P}^3\cong SO(3)$ denote the map which identifies antipodal points of $S^3$. For any $M\in SO(3)$ and $\kappa\in[0,1)$, there exists a symmetric positive definite matrix $\Lambda_{M,\kappa}\in\bbr^{4\times 4}$ such that if $X\sim ACG_3(\Lambda_{M,\kappa})$, then $\Phi(X)\sim \cc_3(M,\kappa)$.
\end{proposition}
For a proof see Appendix \ref{app-cayley}.

When $n=2$ or $n=3$, Eq.(\ref{jacobian}) for the Jacobian of the Cayley transform simplifies to $2^{3n(n-1)/4}(1+\|\phi\|^2)^{1-n}$. One consequence of this fact is that $\cc_n(I_n,\kappa)$ is the pushforward of a Cauchy distribution (multivariate in the case $n=3$) under the Cayley transform.
This fact makes it possible to compute the KL divergence of $\cc_n(M,\kappa)$ from the uniform distribution on $SO(n)$ in these cases. Note that while the prior is uniform on $SO(n)$, reparameterizable sampling is achieved via Cauchy or $ACG_3$ distributions in latent space.
\begin{proposition}\label{kl-prop}
    For $n=2, 3$ the KL divergence of $\cc_n(M,\kappa)$ from the uniform distribution is given by
    \begin{align*}
        \kldiv{\cc_2(M,\kappa)}{U_{SO(2)}} &=-\log(1-\kappa^2)\\
        \kldiv{\cc_3(M,\kappa)}{U_{SO(3)}} &=-\log(1-\kappa^2)-2\log(1-\kappa)-2\kappa.
    \end{align*}
\end{proposition}

For a proof see Appendix \ref{app-kl}.

Thus when learning a distribution of cellular sheaves with special orthogonal restriction maps in our BSNN, for the cases where the stalk dimension $d$ is $2$ or $3$, we are able to use Eq.(\ref{loss_fn}) to estimate the ELBO by plugging in the expressions for the KL divergence from Proposition \ref{kl-prop}. When $d\geq 4$, we estimate the ELBO by simply using Eq.(\ref{loss_fn_est}).
\subsection{Bayesian Sheaf Neural Network Architecture}
We 
present the Bayesian sheaf neural network architecture, which is also summarized in Figure \ref{fig:architecture} and Algorithm \ref{alg:bsnn}. The variational sheaf learning mechanism maps the input node features to the sheaf distributional parameters as described in Section \ref{var_sheaf}, 
For completeness, we briefly review the mechanism here as well. 

A linear layer first resizes the input node features to dimension $k$. Specifically, we use $k=df$ where $d$ and $f$ are the stalk dimensions and number of feature channels respectively. For each edge $e = (u, u') \in E$, we consider the two incident node-edge pairs $u\inc e$ and $u'\inc e$. For each pair, we construct a feature vector by concatenating the features of the node with those of the node at the other end of the edge.
Hence, a node feature matrix of size $n\times k$ is transformed by this concatenation operation into a matrix of size $2e\times 2k$, where $n$ denotes the number of nodes, and $e$ denotes the number of edges in the graph. 

Lastly, an MLP maps these features indexed by the incident node-edge pairs $u\inc e$ to distributional parameters $\mu_{u\inc e}$ and $\sigma_{u\inc e}$ as described by Eq.(\ref{learner}), which define the distribution over the cellular sheaves. From this distribution, a collection of independent and identically distributed sheaves $\mcf_1, \mcf_2, \dots, \mcf_L$ are sampled: one for each layer of the network.

Separately, a multi-layer perceptron maps the input node features $X\in \bbr^{n\times m}$ to features of dimension $X'\in \bbr^{n\times df}$ where $d$ is the stalk dimension and $f$ is the number of hidden channels. The features $X'$ are reshaped to size $nd\times f$.

Sheaf layers described by Eq.(\ref{conv1}) update the transformed features $X'$ using the sampled sheaves $\mcf_1, \mcf_2, \dots, \mcf_L$.
Finally, a linear layer maps the output of the final sheaf layer to a vector of class probabilities.

During training, a one-sample Monte Carlo estimate of Eq.(\ref{loss_fn}) or Eq.(\ref{loss_fn_est}) is computed, and the loss is back-propagated to update the network weights. During testing, multiple sheaf samples are drawn from the learned posterior, and ensembling is performed over the corresponding forward passes to obtain point estimates for the unobserved node labels $y^*$.

\tikzstyle{arrow} = [thick,->,>=stealth]

\tikzstyle{sheaf-layer} = [rectangle, 
minimum width=1cm, 
minimum height=3cm, 
text centered, 
draw=black, 
fill=green!15]

\tikzstyle{input} = [rectangle, 
minimum width=1cm, 
minimum height=3cm, 
text centered, 
draw=black, 
fill=yellow!15]

\tikzstyle{output} = [rectangle, 
minimum width=3cm, 
minimum height=1cm, 
text centered, 
draw=black, 
fill=blue!30]

\tikzstyle{distribution} = [rectangle, 
minimum width=3cm, 
minimum height=1.25cm, 
text centered, 
draw=black, 
fill=blue!15]

\tikzstyle{mlp} = [rectangle, 
minimum width=1cm, 
minimum height=3cm, 
text centered, 
draw=black, 
fill=purple!15]

\tikzstyle{sheaf-mlp} = [rectangle, 
minimum width=2.25cm, 
minimum height=1.5cm, 
text width=2cm,
text centered, 
draw=black, 
fill=purple!15]

\tikzstyle{my-samples} = [rectangle, 
minimum width=1cm, 
minimum height=1cm,
text centered, 
draw=black, 
fill=blue!15]

\begin{figure}[H]
\centering
\resizebox{0.7\textwidth}{!}{
    \begin{tikzpicture}[node distance=2cm]
    
    \node (input) [input] {\rotatebox{90}{Input Graph}};
    \node (mlp1) [mlp, xshift=2cm] {\rotatebox{90}{MLP}};
    
    \node (sheaf1) [sheaf-layer, xshift=4cm] {\rotatebox{90}{Sheaf Conv.}};
    \node (sheaf2) [sheaf-layer, xshift=5.5cm] {\rotatebox{90}{Sheaf Conv.}};
    \node (sheafdots) [xshift=7cm] {$\cdots$};
    \node (sheaf3) [sheaf-layer, xshift=8.5cm] {\rotatebox{90}{Sheaf Conv.}};
    \draw[decoration={brace,raise=1.7cm, amplitude=5pt, mirror},decorate]
      ([xshift=-2pt]sheaf1.west) -- node[above=-2.5cm] {Sheaf Layers} ([xshift=2pt]sheaf3.east);
    
    \node (mlp2) [mlp, xshift=10.5cm] {\rotatebox{90}{Linear Layer}};
    \node (output) [input, xshift=12.5cm] {\rotatebox{90}{Output}};
    
    \node (F1) [my-samples, xshift=4cm, yshift=2.5cm] {$\mcf_1$};
    \node (F2) [my-samples, xshift=5.5cm, yshift=2.5cm] {$\mcf_2$};
    \node (Fdots) [xshift=7cm, yshift=2.5cm] {$\cdots$};
    \node (F3) [my-samples, xshift=8.5cm, yshift=2.5cm] {$\mcf_L$};
    
    \node (dist) [distribution, xshift=6.25cm, yshift=4.75cm] {Sheaf Distribution};
    
    \node (sheaf-mlp) [sheaf-mlp, xshift=2cm, yshift=4.75cm] {Linear Layer + Concat. + MLP};
    
    \draw [arrow] (input) -- (mlp1);
    \draw [arrow] (mlp1) -- (sheaf1);
    \draw [arrow] (sheaf1) -- (sheaf2);
    \draw [arrow] (sheaf2) -- (sheafdots);
    \draw [arrow] (sheafdots) -- (sheaf3);
    
    \draw [arrow] (sheaf3) -- (mlp2);
    \draw [arrow] (mlp2) -- (output);
    
    \draw [arrow] (F1) -- (sheaf1);
    \draw [arrow] (F2) -- (sheaf2);
    \draw [arrow] (F3) -- (sheaf3);
    
    \draw[dashed, arrow] (dist.south) |- ($(dist.south) - (0,0.5cm)$) -| (F1.north);
    \draw[dashed, arrow] (dist.south) |- ($(dist.south) - (0,0.5cm)$) -| (F2.north);
    \draw[dashed, arrow] (dist.south) |- ($(dist.south) - (0,0.5cm)$) -| (F3.north);
    
    \draw [arrow] (input) |- (sheaf-mlp);
    \draw [arrow] (sheaf-mlp) -- (dist);
    
    \end{tikzpicture}
}
\caption{Visual depiction of BSNN Architecture. The variational sheaf learning mechanism is depicted in the upper half of the diagram. Dashed arrows represent sampling from the sheaf distribution.}
\label{fig:architecture}
\end{figure}

\begin{algorithm}[H]
\centering
\footnotesize
\caption{Bayesian Sheaf Neural Network (BSNN)}
\label{alg:bsnn}
\begin{algorithmic}
\REQUIRE Node features $X \in \mathbb{R}^{n \times k}$, graph $G = (V, E)$, stalk dim $d$, hidden channels $f$, layers $L$
\ENSURE Predicted class probabilities for nodes

\STATE $k \gets d \cdot f$
\STATE $X \gets \text{Linear}(X)$ \hfill \COMMENT{Resize to $n \times k$}
\STATE Initialize $H \gets [\,]$ \hfill 

\FOR{each edge $e = (u, u') \in E$}
    \STATE Append $[X_u \| X_{u'}]$ to $H$
    \STATE Append $[X_{u'} \| X_u]$ to $H$
\ENDFOR

\STATE $(\mu, \sigma) \gets \text{MLP}(H)$
\STATE Sample sheaves $F_1, \dots, F_L$
\STATE $X' \gets \text{MLP}(X)$
\STATE Reshape $X'$ to $nd \times f$

\FOR{$\ell = 1$ to $L$}
    \STATE $X' \gets \text{SheafLayer}(X', F_\ell)$
\ENDFOR

\STATE $\hat{y} \gets \text{Linear}(X')$

\IF{training}
    \STATE Compute ELBO and backpropagate
\ELSE
    \STATE Ensemble predictions to estimate $\hat{y}^*$
\ENDIF
\end{algorithmic}
\end{algorithm}

\section{Experiments and Uncertainty Quantification}\label{exp}
We compare our proposed model against a comprehensive set of baseline methods, including classical and advanced graph neural networks (GNNs), and modern deep geometric learning architectures. 
\subsection{Results}
Sheaf neural networks are expected to outperform traditional graph neural networks on datasets with low edge-homophily coefficients, that is, datasets where relatively few edges connect nodes belonging to the same class \citep{zhu2020beyond}. 

We evaluate our model using the NSD benchmark \citep{bodnar2022neural}, which compiles standardized results across datasets with varying levels of homophily, from low (Texas) to high (Cora), enabling fair and diverse comparison. For more details on the datasets, see Appendix~\ref{app:exp-details}.

Baselines include standard GNNs designed for homophilic settings (GCN \citep{kipf2017graph}, GAT \citep{velivckovic2018graph}, GraphSAGE \citep{hamilton2017inductive}), models tailored for heterophilic graphs (GGCN \citep{yan2021two}, Geom-GCN \citep{pei2020geom}, H2GCN \citep{zhu2020beyond}, GPRGNN \citep{chien2021adaptive}, MixHop \citep{abu2019mixhop}), and architectures addressing oversmoothing (GCNII \citep{chen2020simple}, PairNorm \citep{zhao2020pairnorm}). Most results are sourced from \cite{yan2021two}, with MixHop from \cite{bo2021beyond} and \cite{zhu2020beyond}.

\begin{table}[h]
\centering
\footnotesize
\resizebox{\textwidth}{!}{
\begin{tabular}{lccccccc}
\toprule
 & Texas & Wisconsin & Film & Cornell & Citeseer & Pubmed & Cora \\
\midrule
\textbf{Diag-BSNN} & \textbf{\textcolor{purple}{85.95}} $\pm$ 5.77 & 88.43 $\pm$ 2.39 & 37.22 $\pm$ 1.19 & \textbf{\textcolor{orange}{85.95}} $\pm$ 6.71 & 76.91 $\pm$ 1.80 & 89.59 $\pm$ 0.42 & 87.34 $\pm$ 0.94 \\
\textbf{\textit{SO}(\textit{d})-BSNN} & \textbf{\textcolor{purple}{85.95}} $\pm$ 6.26 & \textcolor{blue}{\textbf{89.80}} $\pm$ 3.49 & 37.20 $\pm$ 1.06 & \textcolor{purple}{\textbf{86.22}} $\pm$ 6.10 & 76.80 $\pm$ 1.30 & \textcolor{orange}{\textbf{89.66}} $\pm$ 0.42 & 86.68 $\pm$ 1.14 \\
\textbf{Gen-BSNN} & \textcolor{blue}{\textbf{88.11}} $\pm$ 5.16 & \textbf{\textcolor{purple}{89.61}} $\pm$ 4.21 & 37.31 $\pm$ 0.88 & 85.68 $\pm$ 6.40 & \textbf{\textcolor{orange}{77.17}} $\pm$ 1.84 & 89.63 $\pm$ 0.42 & 87.53 $\pm$ 1.53 \\
\midrule
Diag-NSD & 85.67 $\pm$ 6.95 & 88.63 $\pm$ 2.75 & \textcolor{orange}{\textbf{37.79}} $\pm$ 1.01 & \textcolor{blue}{\textbf{86.49}} $\pm$ 7.35 & 77.14 $\pm$ 1.85 & 89.42 $\pm$ 0.43 & 87.14 $\pm$ 1.06 \\
\textit{O}(\textit{d})-NSD & \textbf{\textcolor{purple}{85.95}} $\pm$ 5.51 & \textcolor{orange}{\textbf{89.41}} $\pm$ 4.74 & \textcolor{blue}{\textbf{37.81}} $\pm$ 1.15 & 84.86 $\pm$ 4.71 & 76.70 $\pm$ 1.57 & 89.49 $\pm$ 0.40 & 86.90 $\pm$ 1.13 \\
Gen-NSD & 82.97 $\pm$ 5.13 & 89.21 $\pm$ 3.84 & \textcolor{purple}{\textbf{37.80}} $\pm$ 1.22 & 85.68 $\pm$ 6.51 & 76.32 $\pm$ 1.65 & 89.33 $\pm$ 0.35 & 87.30 $\pm$ 1.15 \\
GGCN & \textcolor{orange}{\textbf{84.86}} $\pm$ 4.55 & 86.86 $\pm$ 3.29 & 37.54 $\pm$ 1.56 & 85.68 $\pm$ 6.63 & 77.14 $\pm$ 1.45 & 89.15 $\pm$ 0.37 & \textcolor{purple}{\textbf{87.95}} $\pm$ 1.05 \\
H2GCN & \textcolor{orange}{\textbf{84.86}} $\pm$ 7.23 & 87.65 $\pm$ 4.98 & 35.70 $\pm$ 1.00 & 82.70 $\pm$ 5.28 & 77.11 $\pm$ 1.57 & 89.49 $\pm$ 0.38 & \textcolor{orange}{\textbf{87.87}} $\pm$ 1.20 \\
GPRGNN & 78.38 $\pm$ 4.36 & 82.94 $\pm$ 4.21 & 34.63 $\pm$ 1.22 & 80.27 $\pm$ 8.11 & 77.13 $\pm$ 1.67 & 87.54 $\pm$ 0.38 & \textcolor{purple}{\textbf{87.95}} $\pm$ 1.18 \\
MixHop & 77.84 $\pm$ 7.73 & 75.88 $\pm$ 4.90 & 32.22 $\pm$ 2.34 & 73.51 $\pm$ 6.34 & 76.26 $\pm$ 1.33 & 85.31 $\pm$ 0.61 & 87.61 $\pm$ 0.85 \\
GCNII & 77.57 $\pm$ 3.83 & 80.39 $\pm$ 3.40 & 37.44 $\pm$ 1.30 & 77.86 $\pm$ 3.79 & \textcolor{purple}{\textbf{77.33}} $\pm$ 1.48 & \textcolor{blue}{\textbf{90.15}} $\pm$ 0.43 & \textcolor{blue}{\textbf{88.37}} $\pm$ 1.25 \\
Geom-GCN & 66.76 $\pm$ 2.72 & 64.51 $\pm$ 3.66 & 31.59 $\pm$ 1.15 & 60.54 $\pm$ 3.67 & \textcolor{blue}{\textbf{78.02}} $\pm$ 1.15 & \textcolor{purple}{\textbf{89.95}} $\pm$ 0.47 & 85.35 $\pm$ 1.57 \\
PairNorm & 60.27 $\pm$ 4.34 & 48.43 $\pm$ 6.14 & 27.40 $\pm$ 1.24 & 58.92 $\pm$ 3.15 & 73.59 $\pm$ 1.47 & 87.53 $\pm$ 0.44 & 85.79 $\pm$ 1.01 \\
GraphSAGE & 82.43 $\pm$ 6.14 & 81.18 $\pm$ 5.56 & 34.23 $\pm$ 0.99 & 75.95 $\pm$ 5.01 & 76.04 $\pm$ 1.38 & 88.45 $\pm$ 0.50 & 86.90 $\pm$ 1.04 \\
GCN & 55.14 $\pm$ 5.16 & 51.76 $\pm$ 3.06 & 27.32 $\pm$ 1.10 & 60.54 $\pm$ 5.30 & 76.50 $\pm$ 1.36 & 88.42 $\pm$ 0.50 & 86.98 $\pm$ 1.27 \\
GAT & 52.16 $\pm$ 6.63 & 49.41 $\pm$ 4.09 & 27.44 $\pm$ 0.89 & 61.89 $\pm$ 5.05 & 76.55 $\pm$ 1.23 & 87.30 $\pm$ 1.10 & 86.33 $\pm$ 0.48 \\
MLP & 80.81 $\pm$ 4.75 & 85.29 $\pm$ 3.31 & 36.53 $\pm$ 0.70 & 81.89 $\pm$ 6.40 & 74.02 $\pm$ 1.90 & 87.16 $\pm$ 0.37 & 75.69 $\pm$ 2.00 \\
\bottomrule
\end{tabular}}
    \caption{Test accuracies and standard deviations for node classification on datasets sorted in ascending order of homophily. The top three models are colored by rank. Our models are indicated as BSNN.}
    \label{tab:original}
\end{table}

All models, including ours, are evaluated on 10 fixed splits comprising 48\% training, 32\% validation, and 20\% testing data. We report mean accuracy and standard deviation across these splits. Model configurations are selected via random hyperparameter search, with test results reported at the epoch yielding the best validation performance (see Appendix \ref{app:exp-details} Table \ref{tab:config} for the full hyperparameter list). The final model is chosen based on the highest validation score across all candidates. 

The results in Table \ref{tab:original} demonstrate that the BSNN models consistently attain peak or near-peak test accuracies across a wide range of datasets. On heterophilic datasets such as Texas and Wisconsin, BSNN variants are competitive with state-of-the-art methods, with accuracies reaching {88.11\%} on Texas and {89.80\% }on Wisconsin. Similarly, on homophilic benchmarks such as Pubmed and Cora, BSNN variants match or slightly exceed advanced methods, achieving {89.66\%} and {87.53\%} accuracy, respectively. These results indicate that BSNN models are robust to varying degrees of homophily and generalize effectively across different graph structures.

Notably, on the challenging Film dataset with low homophily, sparse connectivity, and high label noise, BSNN models achieve test accuracy comparable to leading methods, indicating robustness to noisy and weakly informative graph structures.

Overall, these results demonstrate that the incorporation of Bayesian uncertainty modeling into graph neural networks provides significant benefits across a broad spectrum of datasets.

\subsection{Uncertainty Quantification}
Given that BSNN models learn predictive distributions on sheaves, we evaluate their reliability, and calibration using metrics such as entropy, epistemic variance, mutual information, and expected calibration error as outlined in Section \ref{uq_metrics}. 
We conduct independent training trials using 30 random seeds with the best hyperparameter configuration, namely the one that achieved the highest validation accuracy. For each trial, a BSNN model samples a new sheaf instance and performs three forward passes. We use these predictions to compute node-wise uncertainty statistics.

\begin{figure}
    \centering
    \includegraphics[width=1\linewidth]{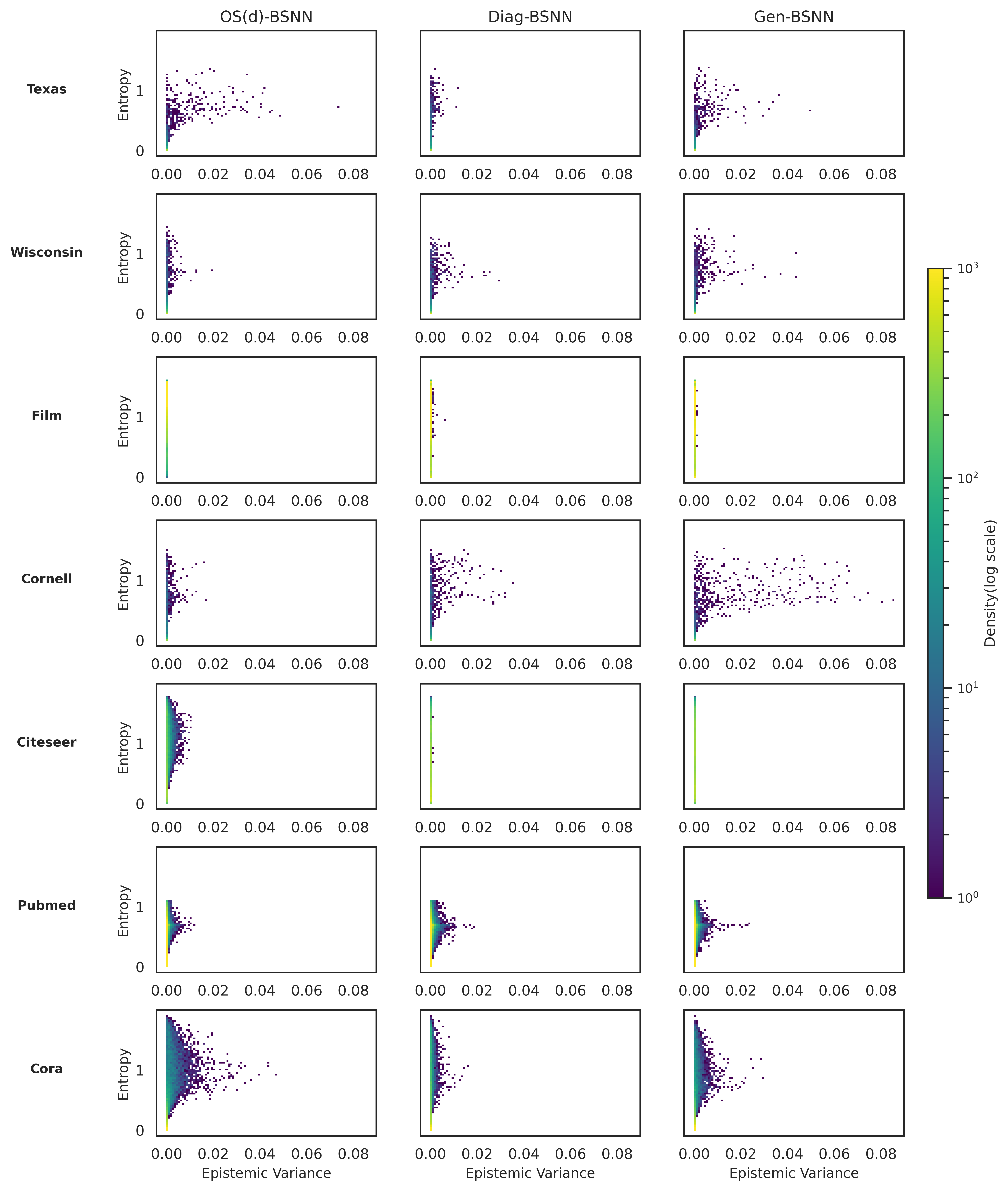}
    \caption{Predictive entropy and epistemic variance for BSNN models, illustrating total and model uncertainty across nodes.}
    \label{fig:entropy}
\end{figure}

For each test node, we calculate entropy to quantify \emph{total uncertainty}, and estimate epistemic variance to capture \emph{model uncertainty}. As shown in Figure \ref{fig:entropy}, epistemic variance is more sensitive to model choice than entropy. This variability in epistemic uncertainty across models is natural and expected, as different models embody distinct prior beliefs and architectures, which interact with datasets in different ways. In contrast, entropy remains relatively stable across models for a given dataset, suggesting that it is likely dominated by aleatoric uncertainty.

Most nodes exhibit low epistemic variance (near 0) and low to moderate entropy, suggesting that models are generally confident and consistent in their predictions, with no clear correlation between epistemic variance and homophily level across models.
The maximum observed epistemic variance (0.08) is  on Cornell when Gen-BSNN is used, while the rest of the plots show a maximum epistemic variance of about 0.04. 
Node degree does not appear to significantly influence epistemic variance, as Citeseer and Film both exhibit low variance under Gen-BSNN despite having markedly different average degrees, where Citeseer being sparsely connected and Film more densely connected. 

Among the three BSNN variants, Diag-BSNN exhibits lower epistemic variance than Gen-BSNN and $SO(d)$-BSNN in most datasets. This suggests that its simplified diagonal posterior often leads to more conservative uncertainty estimates, but its behavior still varies depending on the specific dataset characteristics.

\begin{figure}
    \centering
    \includegraphics[width=0.9\linewidth]{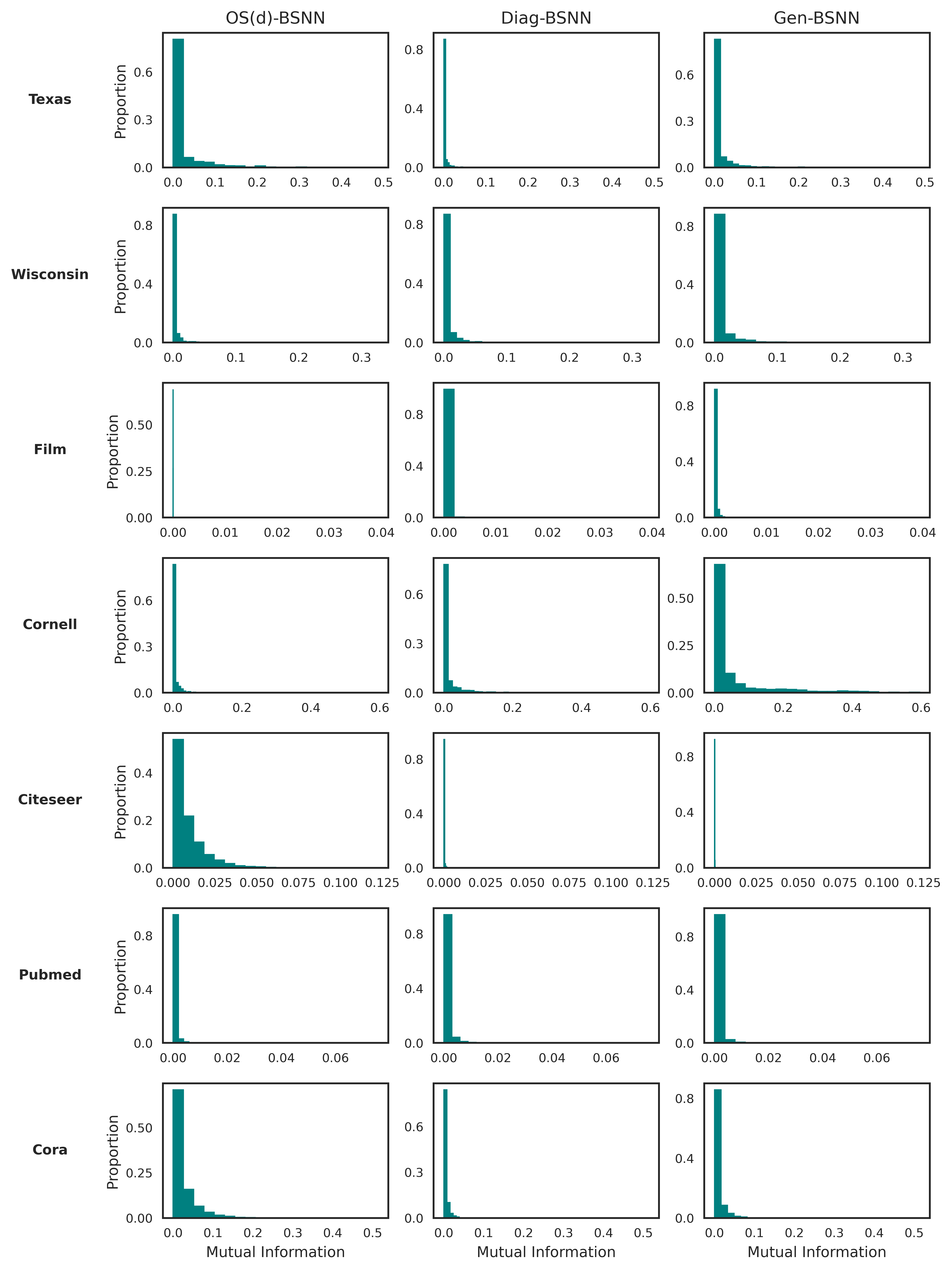}
    \caption{Mutual information for BSNN models as a measure of epistemic uncertainty.}
    \label{fig:mi}
\end{figure}

In most datasets, models demonstrate confident predictions for the majority of nodes, as indicated by a high concentration of low-entropy values. Only a smaller subset of nodes shows elevated entropy and variability, which typically points to localized uncertainty arising from ambiguous features. In contrast, the Film dataset shows more consistent entropy levels across nodes, with overall low variance, suggesting that the observed uncertainty is likely due to noise rather than informative structure. This interpretation is supported by the low mutual information (MI) values, as will be shown in Figure \ref{fig:mi}.

Table \ref{tab:average}  presents average uncertainty scores, computed by first averaging predictive entropy and epistemic variance within each dataset, and then averaging those values across all datasets. The observed averages suggest that all models are similarly confident with moderately low uncertainty overall. 

\vspace{1em}
\begin{table}[h]
\centering
\resizebox{0.6\textwidth}{!}{
\begin{tabular}{lccc}
\toprule
& Average Entropy & Average Epistemic Variance \\
\midrule
\textbf{Diag-BSNN} & 0.474796 & 0.000381 \\
\textbf{Gen-BSNN} & 0.449049 & 0.001247 \\
\textbf{\textit{SO}(\textit{d})-BSNN} & 0.525914 & 0.000738 \\
\bottomrule
\end{tabular}}
\caption{Average entropy and epistemic variance for BSNN models across all datasets.}
\label{tab:average}
\end{table}


Since predictive entropy is nearly invariant across models, we compute mutual information (MI) to better capture differences in uncertainty. The histograms that illustrate mutual information in Figure \ref{fig:mi} are heavily skewed toward 1.0 across all datasets and all models. This suggests that the BSNN models are very confident about most of their predictions. As for Cornell, a small proportion of samples have a higher MI, meaning the model is uncertain about these specific samples, likely due to out-of-distribution features, or decision boundary proximity. This is not surprising, as these samples also exhibited the highest epistemic uncertainty, further indicating the model's lack of confidence in its predictions for these cases.

We compute the Expected Calibration Error (ECE) to assess the alignment between predicted confidence and actual accuracy, and report the results in Table \ref{tab:ece}. On datasets with high homophily levels, the ECE values that are below 0.1 show that the predicted probabilities closely align with actual outcomes, and that the models are well-calibrated. In contrast, on heterophilic datasets such as \textit{Texas}, \textit{Wisconsin}, and \textit{Film}, the ECE values are higher. This suggests that while the models may achieve high accuracy, they are overconfident. To address this, post-hoc calibration techniques such as temperature scaling could be applied to improve the reliability of confidence scores without retraining the model.
\begin{table}[h]
\centering
\resizebox{0.8\textwidth}{!}{
\begin{tabular}{lccccccc}
\toprule
 & Texas & Wisconsin & Film & Cornell & Citeseer & Pubmed & Cora \\
\midrule
\textbf{Diag-BSNN} &  0.1483& \textcolor{blue}{0.0918 }& 0.2490 & 0.1564 & \textcolor{blue}{0.0391} & \textcolor{blue}{0.0113} & \textcolor{blue}{0.0473} \\
\textbf{\textit{SO}(\textit{d})-BSNN} & 0.2061 & 0.1128 & 0.1168 & 0.2602 & \textcolor{blue}{0.0401} & \textcolor{blue}{0.0144} & \textcolor{blue}{0.0296} \\
\textbf{Gen-BSNN} & 0.1998 & 0.1343 & 0.2652 & 0.2072 & \textcolor{blue}{0.0440} & \textcolor{blue}{0.0144} & \textcolor{blue}{0.0449} \\
\bottomrule
\end{tabular}}
\caption{Expected Calibration Error (ECE) for different BSNN variants across datasets. Values below 0.1 are highlighted in blue to indicate well-calibrated models.}
\label{tab:ece}
\end{table}
\subsection{Robustness Under Limited Data}
To evaluate the performance of our Bayesian sheaf models under limited training data, we use 10 fixed dataset splits, with each split allocating 32\% of the nodes for training, 20\% for validation, and 48\% for testing. For each model configuration, we report test results from the epoch with the best validation score. The final model is selected as the configuration with the highest validation performance overall. As shown in Table \ref{tab:less}, our model still achieves high accuracy in most datasets, even in this low-data regime.
\begin{table}[h]
\centering
\resizebox{\textwidth}{!}{
\begin{tabular}{lccccccc}
\toprule
 & Texas & Wisconsin & Film & Cornell & Citeseer & Pubmed & Cora \\
\midrule
\textbf{Diag-BSNN} & 80.11 ± 5.24 & 83.25 ± 5.16 & 36.83 ± 0.74 & 74.60 ± 3.19 & 73.71 ± 1.04 & 88.74 ± 0.26 & 84.52 ± 1.25 \\
\textbf{\textit{SO}(\textit{d})-BSNN} & 79.31 ± 7.27 & 84.00 ± 3.96 & 35.47 ± 0.78 & 74.37 ± 4.99 & 73.95 ± 2.59 & 88.92 ± 0.29 & 83.98 ± 1.06 \\
\textbf{Gen-BSNN }& 79.66 ± 5.34 & 83.33 ± 4.62 & 36.83 ± 0.56 & 74.25 ± 5.22 & 73.98 ± 1.18 & 88.72 ± 0.36 & 84.50 ± 1.12 \\
\bottomrule
\end{tabular}}
\caption{Test accuracies and standard deviations for node classification under limited training data conditions, using the hyperparameter configuration yielding the best validation accuracy.}
\label{tab:less}
\end{table}

Additionally, to examine the sensitivity of the Bayesian and deterministic sheaf models to the hyperparameter selections, we perform a grid search over selected hyperparameters, as detailed in Appendix \ref{app:exp-details}, Table~\ref{tab:hyp}. We report the mean test accuracy across \emph{all hyperparameter configurations} in Table \ref{tab:less_hyp}. 

The advantage of each BSNN model over the corresponding NSD model when considering the entire range of hyperparameters is definitive. 
A Wilcoxon signed-rank test for a $p$-value threshold of $p=0.05$ indicates a statistically significant improvement in performance for the Diag-BSNN and Gen-BSNN models over their NSD counterparts across datasets. 

Thus, we observe that the BSNN models maintain good performance over a much wider range of hyperparameters when compared with the NSD model with the same type of restriction maps. We also include results for a graph convolutional network (GCN) as a baseline comparison.
\begin{table}[ht]
\centering
\resizebox{\textwidth}{!}{
\begin{tabular}{lccccccc}
\toprule
 & Texas & Wisconsin & Film & Cornell & Citeseer & Pubmed & Cora \\
\midrule
\textbf{Diag-BSNN} & \textcolor{purple}{\textbf{73.48}} ± 2.65 & \textcolor{purple}{\textbf{77.56}} ± 3.60 & \textcolor{purple}{\textbf{35.36}} ± 1.02 & \textcolor{purple}{\textbf{65.17}} ± 5.71 & \textcolor{purple}{\textbf{68.53}} ± 3.97 & 87.76 ± 0.59 & \textcolor{purple}{\textbf{76.14}} ± 5.13 \\
\textbf{\textit{SO}(\textit{d})-BSNN} & 72.75 ± 3.51 & 76.96 ± 4.59 & 34.51 ± 1.26 & 64.90 ± 6.69 & 68.34 ± 4.91 & 87.88 ± 0.63 & 76.09 ± 5.40 \\
\textbf{Gen-BSNN} & \textcolor{blue}{\textbf{73.73}} ± 2.67 & \textcolor{blue}{\textbf{78.05}} ± 3.25 & \textcolor{blue}{\textbf{35.43}} ± 0.85 & \textcolor{blue}{\textbf{66.35}} ± 4.53 & \textcolor{blue}{\textbf{69.08}} ± 3.56 & 87.81 ± 0.53 & 76.63 ± 4.59 \\
\midrule
Diag-NSD & 71.22 ± 3.31 & 74.91 ± 4.69 & 34.99 ± 1.31 & 62.43 ± 7.11 & 66.59 ± 5.76 & \textcolor{purple}{\textbf{88.01}} ± 0.58 & 75.26 ± 7.85 \\
\textit{O}(\textit{d})-NSD & 71.86 ± 3.16 & 74.63 ± 5.07 & 34.63 ± 1.26 & 61.87 ± 6.38 & 65.14 ± 6.37 & \textcolor{blue}{\textbf{88.03}} ± 0.63 & 73.04 ± 9.65 \\
Gen-NSD & 71.44 ± 2.47 & 74.24 ± 3.91 & 35.11 ± 1.00 & 61.93 ± 5.79 & 64.17 ± 6.85 & 87.87 ± 0.48 & 73.48 ± 10.5 \\
\midrule
GCN & 55.11 ± 1.42 & 50.23 ± 2.81 & 27.22 ± 1.77 & 52.27 ± 0.39 & 70.19 ± 2.22 & 86.01 ± 0.99 & \textcolor{blue}{\textbf{84.21}} ± 1.24 \\
\bottomrule
\end{tabular}}
\caption{Average test accuracy and corresponding standard deviations across all hyperparameter configurations, evaluated under limited training data conditions.}
\label{tab:less_hyp}
\end{table}
%
%
\vspace{-1em}
\section{Conclusion and Future Directions}\label{discussion}

In this work, we have introduced Bayesian sheaf neural networks: novel sheaf neural networks in which the sheaf Laplacian is a random latent variable, and where the loss function used in training contains a Kullback-Leibler divergence regularization term. Not only does this allow one to quantify uncertainty of the model arising from the learned sheaf, but we show that
our BSNN models 
achieve results that are comparable to, and in some cases exceed, state-of-the-art node classification methods.
Furthermore, our experiments suggest that the BSNN models are more robust to different choices of hyperparameters and weight initializations, particularly under limited training data, compared to their deterministic counterparts.

We hope that this paper encourages further work on stochastic sheaf neural networks. We briefly mention some possible future directions. Different priors for the distribution of sheaf Laplacians could be suitable for the BSNN depending on the graph dataset, in particular, depending on the level of edge homophily. Although we view the underlying graph as fixed in this paper, one could view it probabilistically and combine our work with the approach taken in \cite{zhang2019bayesian}. Lastly, similar to the continuous-time version of the sheaf diffusion network that was studied in \cite{bodnar2022neural}, a continuous-time version of the BSNN could be considered, where the sheaf Laplacian is a stochastic process.

\section*{Acknowledgments}
This work has been partially supported by the Army Research Laboratory Cooperative Agreement No W911NF2120186, and STRONG ARL CA No W911NF-22-2-0139.

\bibliography{references}
\clearpage
\appendix
\appsection{Proof of Proposition \ref{sheaf-power}}\label{app-sheaf-power}

In what follows, given a cellular sheaf $\mcf$ over a graph $G=(V,E)$ and initial conditions $X(0)$, let $X(t)$ denote the solution to Eq.(\ref{ode}).
Note that the limit of $X(t)$ as $t\to \infty$ exists, and is equal to the orthogonal projection of $X(0)$ onto $\ker(\Delta_\mcf)$ \citep{hansen2021opinion}. If $\mcf$ has $d${\dash}dimensional stalks and $G$ has $n$ nodes, although $X(t)\in\bbr^{nd}$, it will be useful to equally regard $X(t)$ as an $n\times d$ matrix. 

Let $X(t)_u$ denote the row of $X(t)$ corresponding to node $u$. We write $u\in V_l$ if node $u$ has label $l$. For a cellular sheaf $\mcf$ with $d${\dash}dimensional stalks over a graph $G=(V, E)$ with labeling $\ell:V\to\mcl$ and initial conditions $X(0)$, we say that $\mcf$ \emph{linearly separates the classes of $(G, X(0), \ell)$} if for each label $l\in\mcl$, the sets $\{\lim_{t\to\infty}X(t)_u:u\in V_l\}\subseteq\bbr^d$ and $\{\lim_{t\to\infty}X(t)_u:u\notin V_l\}\subseteq\bbr^d$ are linearly separable. 

With this terminology, Definition \ref{power} states that a collection of cellular sheaves $\mch$ has linear separation power over a collection of labeled graphs $\mcg$, if for each graph $G=(V,E)\in\mcg$ with labeling $\ell:V\to \mcl$, there exists a cellular sheaf $\mcf\in \mch$ over $G$ such that $\mcf$ linearly separates the classes of $(G, X(0), \ell)$ for almost all initial conditions $X(0)$, i.e. all but a measure zero subset of the initial conditions.

\begin{proof}[\textbf{Proof of Proposition \ref{sheaf-power}}]
    Let $d\geq 1$ and let $d=4k+m$ for $0\leq m<4$. That is, $k=\lfloor\tfrac{d}{4}\rfloor$. Since proving the proposition for $C=7k$ also proves the cases $C<7k$, we may assume that $C=7k$. Let $\mcl=\{1, 2, \dots, C\}$ denote the set of class labels for the nodes of a graph $G=(V,E)\in\mcg$ and let $\ell:V\to \mcl$ denote the labeling of the nodes of $G$. 
    
    For each $1\leq i\leq k$, let $J_i=\{7(i-1)+1, 7(i-1)+2,\dots,7(i-1)+7\}$, let $\mcl_i=\{j:j\in J_i\}\cup \{0\}$, and let $\ell_i:V\to\mcl_i$ denote a relabeling of $G$ where for each $u\in V$, $f_i(u)=f(u)$ whenever $f(u)\in J_i$, and $f_i(u)=0$ otherwise. That is, the relabeling $\ell_i$ retains the seven labels in $J_i$, and replaces all other labels with a single distinct eighth label.

    According to Proposition 13 in \cite{bodnar2022neural}, for each $1\leq i\leq k$ we may find a cellular sheaf $\mcf^i\in \mathcal{H}_{so}^4$ with $4${\dash}dimensional stalks which linearly separates the eight classes of $(G, X(0), \ell_i)$ for almost all initial conditions $X(0)\in\bbr^{4n}$. Let $\mcf$ denote the cellular sheaf with restriction maps $\mcf_{u\inc e}=\mcf^1_{u\inc e}\oplus \mcf^2_{u\inc e}\oplus\cdots\oplus\mcf^k_{u\inc e}\oplus I_m$. We claim that $\mcf$ linearly separates the classes of the original labeling $(G, X(0), \ell)$ for almost all initial conditions $X(0)\in \bbr^{n\times d}$. 

    Given initial conditions $X(0)\in \bbr^{n\times d}$, for each $1\leq i\leq k$ let $X_i(0)\in\bbr^{n\times 4}$ be the matrix consisting of columns $4i-3$ through $4i$ of $X(0)$. Note that for almost all initial conditions $X(0)$, it will be true that for all $1\leq i\leq k$, $\mcf^i$ linearly separates the classes of $(G, X_i(0),\ell_i)$. Next, for any class $l\in \mcl$, find $i$ such that $l\in J_i$. Since $\mcf^i$ linearly separates the classes of $(G, X_i(0),\ell_i)$, we may find a hyperplane $A\subseteq \bbr^4$ which separates the class $l$ from all other classes in $\mcl_i$, and hence $\mcl$ as well. Then $\bbr^4\oplus\dots\oplus A\oplus\dots\oplus\bbr^4\oplus\bbr^m$ is a hyperplane in $\bbr^d$ separating class $l$ from all other classes.
\end{proof}
\noindent 
\vspace{-0.5em}
\appsection{Section \ref{cayley_dist} Proofs}\label{app-cayley}
For a subset $S$ of a metric space $M$ and for $\delta>0$, let
\begin{equation*}
    \mch^d_\delta(S)=\inf\Big\{\sum_{i=1}^\infty \alpha_d (\tfrac{1}{2}\diam(S))^d:\bigcup_{i=1}^\infty U_i\supseteq S, \diam(U_i)<\delta\Big\},
\end{equation*}
where $\alpha_d$ denotes the volume of a unit $d${\dash}ball. The $d${\dash}dimensional Hausdorff measure on $M$ is then given by $\mch^d(S)=\lim_{\delta\to 0}H^d_\delta(S)$.

Let $\lambda^n$ denote the $n${\dash}dimensional Lebesgue measure. The area formula in \cite{federer} computes the Hausdorff measure of the image of a $1$-to-$1$ and continuously differentiable function $f$ in terms of the Jacobian of $f$. We state a version presented in \cite{traynor}.

\begin{proposition}[Area formula]\label{area-formula}
    Let $f:\bbr^n\to \bbr^m$ be $1$-to-$1$ and continuously differentiable. For any Borel subset $A\subseteq \bbr^n$,
    \begin{equation*}
    \int_A Jf(x)\lambda^n(dx)=\int_{f(A)}\mch^n(dy).
    \end{equation*}
    Moreover,
    \begin{equation}\label{change-of-var}
    \int_A g(f(x))Jf(x)\lambda^n(dx)=\int_{f(A)}g(y)\mch^n(dy), 
    \end{equation}
    for all Borel measurable $g:\bbr^m\to\bbr$ for which one side of (\ref{change-of-var}) exists.
\end{proposition}

Equip $SO(n)$ with the metric induced by the Frobenius norm $\|A\|=\sqrt{\tr(A^\T A)}$. Since left (or right) multiplication on $SO(n)$ is an isometry, it follows that the Hausdorff measure is a left (and right) translation-invariant measure on $SO(n)$. Moreover, if $d=n(n-1)/2$ denotes the dimension of $SO(n)$, then $\mch^d$ is a locally finite measure on $SO(n)$, and thus is a Haar measure by Haar's theorem.

Hence, Proposition \ref{area-formula} yields a change of variables formula for the Cayley transform. For example, if $X$ is a uniformly distributed random variable on $SO(n)$, then $C^{-1}(X)$ has density $p(x)=J\wt C(x)$ with respect to the Lebesgue measure on $\bbr^d$.

\begin{proof}[\textbf{Proof of Theorem \ref{cayley-thm}}]
Let $\mu$ denote the unique Haar measure on $SO(n)$ such that $\int_{SO(n)}d\mu = 1$. Let $X$ be a uniformly distributed random variable on $SO(n)$. Let $Y=C(\frac{1-\kappa}{1+\kappa} C^{-1}(X))$, and let $f_n(P;I,\kappa)$ denote the density of $Y$ with respect to $\mu$. For convenience, set $\gamma=\frac{1+\kappa}{1-\kappa}$. Then by Proposition \ref{area-formula} and the chain rule,
\begin{equation*}
    f_n(P;I, \kappa)=\gamma^{\frac{n(n-1)}{2}}J\wt C\Big(\gamma\wt C^{-1}(\vect P)\Big)J\wt C^{-1}(\vect P).
\end{equation*}
Using Eq.(\ref{jacobian}), we obtain
\begin{equation}\label{density-eq}
    f_n(P;I, \kappa)=\gamma^{\frac{n(n-1)}{2}} \Big(\frac{\det(I+C^{-1}(P))}{\det(I+\gamma C^{-1}(P))}\Big)^{n-1}.
\end{equation}
Using the equality $\det(I+C^{-1}(P))=\frac{2^n}{\det(I+P)}$,
\begin{equation*}
    f_n(P;I, \kappa)=2^{n(n-1)}\gamma^{\frac{n(n-1)}{2}} \det\Big((I+P)(I+\gamma C^{-1}(P))\Big)^{1-n}.
\end{equation*}
Next, from Eq.(\ref{cayley_eq}) we have that $I+\gamma C^{-1}(P)=(1+\gamma)I - 2\gamma (I+P)^{-1}$. By substituting and simplifying,
\begin{equation*}
    f_n(P;I, \kappa)=2^{n(n-1)}\gamma^{\frac{n(n-1)}{2}} \det\Big((1+\gamma)(I+P) - 2\gamma I\Big)^{1-n}.
\end{equation*}
Factoring $1+\gamma$ out of the expression within the determinant, we have
\begin{equation*}
    f_n(P;I, \kappa)=2^{n(n-1)}\Big(\frac{\gamma}{(1+\gamma)^2}\Big)^{\frac{n(n-1)}{2}} \det\Big(P -\frac{\gamma-1}{\gamma+1} I\Big)^{1-n},
\end{equation*}
or in terms of $\kappa$,
\begin{equation*}
    f_n(P;I, \kappa)=(1-\kappa^2)^{\frac{n(n-1)}{2}} \det\Big(P -\kappa I\Big)^{1-n}.
\end{equation*}
\end{proof}

\begin{proof}[\textbf{Proof of Proposition \ref{acg-prop}}]
    Representing the points of $S^3$ as unit vectors $x=(a,b,c,d)\in \bbr^4$, the $2${\dash}fold cover $\Phi:S^3\to SO(3)$ has an explicit description as
\begin{equation*}
    \Phi(x)
=\begin{pmatrix}
    a^2+b^2-c^2-d^2 & 2(bc-ad) & 2(bd+ac) \\
    2(bc+ad) & a^2-b^2+c^2-d^2 & 2(cd-ab) \\
    2(bd-ac) & 2(cd+ab) & a^2-b^2-c^2+d^2
\end{pmatrix}.
\end{equation*}

A straightforward calculation shows that
\begin{equation*}
    \det(\Phi(x)-\kappa I)=(1-\kappa)^3a^2+(1+\kappa)^2(1-\kappa)(b^2+c^2+d^2).
\end{equation*}
If $f(P;M, \kappa)$ denotes the density of $\cc_3(M,\kappa)$ with respect to the unit Haar measure on $SO(3)$, $g(x;\Lambda)$ denotes the density of $ACG_3(\Lambda)$ with respect to the unit Haar measure on $S^3$, and $\Lambda_\kappa$ is the diagonal matrix $\Lambda_\kappa=\diag((\frac{1+\kappa}{1-\kappa})^2, 1, 1, 1)$, then we have
\begin{align*}
   f(\Phi(x);I, \kappa) &\propto \det(\Phi(x)-\kappa I)^{-2}\\ 
   & = \Big((1-\kappa)^3a^2+(1+\kappa)^2(1-\kappa)(b^2+c^2+d^2)\Big)^{-2}\\
   & \propto \Big(\big(\frac{1-\kappa}{1+\kappa}\big)^2a^2 + b^2 + c^2 +d^2\Big)^{-2}\\
   & \propto g(x;\Lambda_\kappa),
\end{align*}
where the symbol $\propto$ denotes "proportional to".

Moreover, for $M\in SO(3)$, let $y_M\in \Phi^{-1}(M^\T)$. Regarding $y_M$ as a unit quaternion, let $Q\in SO(4)$ be the rotation $z\mapsto zy_M$ for $z\in\mathbb{H}$. Note that $\Phi(x)M^\T=\Phi(xy_M)=\Phi(Qx)$ and $g(x;Q^\T \Lambda_\kappa Q)=g(Qx;\Lambda_\kappa)$. Hence 
\begin{equation*}
f(\Phi(x);M,\kappa)=f(\Phi(Qx);I,\kappa)\propto g(x;Q^\T \Lambda_\kappa Q).
\end{equation*}

Lastly, since $\Phi$ is a homomorphism, the pushforward of the unit Haar measure on $S^3$ yields the unit Haar measure on $SO(3)$. This combined with the fact that $f(\Phi(x);M,\kappa)\propto g(x;Q^\T \Lambda_\kappa Q)$ implies that $\Phi(X)\sim \cc_3(M,\kappa)$ whenever $X\sim ACG_3(\Lambda_{M,\kappa})$ for $\Lambda_{M,\kappa}=Q^\T \Lambda_\kappa Q$.
\end{proof}

\appsection{KL Divergence Calculations}\label{app-kl}
Recall that we assume that the variational distribution $q_\phi(\mcf|X,y)$ factors as $q_\phi(\mcf|X,y)=\prod_{u\inc e}q_\phi(\mcf_{u\inc e}|X,y)$ over all incident node-edge pairs $u\inc e$, and that we use a prior $p_\theta(\mcf)$ which also factors in the same manner $p_\theta(\mcf)=\prod_{u\inc e}p_\theta(\mcf_{u\inc e})$. As a consequence of the chain rule for KL divergence, see for example Theorem 2.5.3 in \cite{cover2012elements}, we have
$$\kldiv{q_\phi(\mcf|X,y)}{p_\theta(\mcf)}=\sum_{u\inc e} \kldiv{q_\phi(\mcf_{u\inc e}|X,y)}{p_\theta(\mcf_{u\inc e})}.$$
Hence it suffices to be able to compute $\kldiv{q_\phi(\mcf_{u\inc e}|X,y)}{p_\theta(\mcf_{u\inc e})}$ for all $u\inc e$. 

In the cases where 
the sheaf restriction maps are diagonal or general linear, this divergence reduces to the KL between a Gaussian with diagonal covariance and a standard normal distribution.
The KL divergence of a normal distribution $\mathcal{N}(\mu, \diag(\sigma^2))$ with mean $\mu\in\bbr^d$ and $\sigma^2\in\bbr^d$ from the standard normal distribution is
\begin{equation}\label{normal-kl}
    \kldiv{\mathcal{N}(\mu, \sigma^2)}{\mathcal{N}(0, 1)} = -\frac{1}{2}\sum_{i=1}^d\big(1+\log(\sigma^2_i)-\mu_i^2-\sigma_i^2\big).
\end{equation}

In the case where the restriction maps are special orthogonal, and the stalk dimension $d$ is $2$ or $3$, we compute $\kldiv{q_\phi(\mcf_{u\inc e}|X,y)}{p_\theta(\mcf_{u\inc e})}$ using Proposition \ref{kl-prop}, which we now prove.
\begin{proof}[\textbf{Proof of Proposition \ref{kl-prop}}]
    First, we note that the integral of a function $f(P)$ on $SO(n)$ with respect to a Haar measure is invariant under the change of variables $P\mapsto PM$ for any $M\in SO(n)$.
    Hence $\kldiv{\cc_n(M,\kappa)}{U_{SO(n)}}$ does not depend on $M$, that is,
    \begin{equation*}
        \kldiv{\cc_n(M,\kappa)}{U_{SO(n)}}=\kldiv{\cc_n(I,\kappa)}{U_{SO(n)}}.
    \end{equation*}
    To simplify notation, set $d=n(n-1)/2$, let $\gamma=\frac{1+\kappa}{1-\kappa}$, and let $V_n$ denote the volume of $SO(n)$. Recall Eq.(\ref{density-eq}), and note that the density of $\cc_n(I, \kappa)$ with respect to $\mch^{d}$ can be expressed as 
    \begin{equation*}
        V_n^{-1}\gamma^d \Big(\frac{\det(I+C^{-1}(P)}{\det(I+\gamma C^{-1}(P))}\Big)^{n-1},
    \end{equation*}
    where $V_n$ denotes the volume of $SO(n)$. Then the change of variables formula  Eq.(\ref{change-of-var}) yields
    \begin{equation*}
        \kldiv{\cc_n(M,\kappa)}{U_{SO(n)}}= \int_{\bbr^d} 
    \frac{2^{\frac{3n(n-1)}{4}}V_n^{-1}\gamma^d}{\det(I+\gamma X_\phi)^{n-1}} \log\Big( \frac{\gamma^d\det(I+X_\phi)^{n-1}}{\det(I+\gamma X_\phi)^{n-1}}\Big)\, \lambda^d(d\phi),
    \end{equation*}
    where recall from Section \ref{cayley_dist} that $X_\phi$ denotes the skew symmetric matrix whose entries below the diagonal are the entries of the vector $\phi$.

    When $n=2, 3$, we have $\det(I+X_\phi)=1+\lVert \phi \rVert^2$ and $\det(I+\gamma X_\phi)=1+\gamma^2\lVert \phi \rVert^2$, hence
    \begin{equation*}
        \kldiv{\cc_n(M,\kappa)}{U_{SO(n)}}= \int_{\bbr^d}
    \frac{2^{\frac{3n(n-1)}{4}}V_n^{-1}\gamma^d}{(1+\gamma^2\lVert \phi \rVert^2)^{n-1}} \log\Big(\frac{\gamma^d(1+\lVert \phi \rVert^2)^{n-1}}{(1+\gamma^2\lVert \phi \rVert^2)^{n-1}}\Big)\, \lambda^d(d\phi).
    \end{equation*}

    Recall that by \citep{ponting1949volume},
    \begin{equation*}
        V_n=2^{\frac{3n(n-1)}{4}}\prod_{i=2}^n \frac{\pi^{(i-1)/2}\Gamma((i-1)/2)}{\Gamma(i-1)}.
    \end{equation*}

    Hence $\kldiv{\cc_n(M,\kappa)}{U_{SO(n)}}$ is precisely the KL divergence of $\Cauchy_d(0, \frac{1}{\gamma^2}I)$ from $\Cauchy_d(0, I)$ for $n=2$ or $n=3$. By the expressions for the KL divergence between Cauchy distributions in \cite{e24060838}, we have
    \begin{equation*}
        \kldiv{\Cauchy_1(0, \tfrac{1}{\gamma^2}I)}{\Cauchy_1(0, I)}=\log\bigg(\frac{(\gamma+1)^2}{4\gamma}\bigg),
    \end{equation*}
    and
    \begin{equation*}
        \kldiv{\Cauchy_3(0, \tfrac{1}{\gamma^2}I_3)}{\Cauchy_3(0, I_3)}=\log\bigg(\frac{(\gamma+1)^4}{16\gamma}\bigg)+\frac{2(1-\gamma)}{1+\gamma}.
    \end{equation*}
    Since $\gamma=\frac{1+\kappa}{1-\kappa}$
    we have
    \begin{equation*}
        \kldiv{\cc_2(M,\kappa)}{U_{SO(2)}}=-\log(1-\kappa^2),
    \end{equation*}
    and
    \begin{equation*}
        \kldiv{\cc_3(M,\kappa)}{U_{SO(3)}}=-\log(1-\kappa^2)-2\log(1-\kappa)-2\kappa.
    \end{equation*}
\end{proof}
\noindent 
\vspace{-0.5em}
\appsection{Experiment Details}\label{app:exp-details}
We selected a diverse set of real-world datasets exhibiting varying levels of homophily. Table \ref{tab:dataset} shows the number of nodes, edges, and classes in each dataset. 

The Texas, Cornell, and Wisconsin datasets are part of the WebKB collection, each representing the computer science department of a respective university. Each dataset is a graph in which nodes represent web pages and edges represent hyperlinks between them. In all datasets the node features are $1703${\dash}dimensional bag-of-word representations of the web pages.

The Film dataset is a graph of actors, where nodes represent actors and edges represent co-appearances in films. Node features are derived from metadata such as keywords and genres, and each node belongs to one of several genre-based classes. 

The Pubmed dataset is a citation network of scientific publications. Each node represents a paper, and edges represent citation links. Node features are TF/IDF (Term Frequency-Inverse Document Frequency) weighted word vectors from the paper abstracts, and nodes are labeled according to their subject area. 

The CiteSeer and Cora datasets are also citation networks. Nodes represent documents, and edges indicate citations between them. Node features are bag-of-word vectors, and each document is categorized into a specific topic. 

\begin{table}[h!]
\centering
\begin{tabular}{l|rrrr}

\textbf{Dataset} & \textbf{Nodes} & \textbf{Edges} & \textbf{Classes}  \\
\hline
Texas     & 183    & 279    & 5  \\
Wisconsin & 251    & 450    & 5 \\
Film      & 7600   & 26659  & 5  \\
Cornell   & 183    & 277    & 5  \\
Citeseer  & 3327   & 4552   & 6  \\
Pubmed    & 19717  & 44324  & 3 \\
Cora      & 2708   & 5278   & 7\\
\hline
\end{tabular}
\caption{Dataset statistics, including the number of nodes, edges, and classes.}
\label{tab:dataset}
\end{table}

Chameleon and Squirrel, which were included in earlier studies, were excluded from our experiments because they have been removed from the official Geom-GCN repository due to maintenance or quality concerns.

All experiments were run on a computing cluster equipped with an NVIDIA H100 GPU (80GB HBM3), two Intel Xeon Platinum 8462Y+ CPUs, and 1.0 TiB of RAM. The code was implemented in PyTorch and executed using CUDA 12.4 with NVIDIA driver 550.54.15.. Our code is available at \url{https://github.com/Layal-Bou-Hamdan/BSNN}.

Hyperparameter tuning was done using Weights \& Biases \citep{wandb}. The full range of hyperparameters that we optimize using random search is listed in Table \ref{tab:config}.

\begin{table}[htbp]
\centering
\begin{tabular}{@{}ll@{}}
\toprule
\textbf{Hyperparameter} & \textbf{Values} \\
\midrule
Hidden channels         & \{8, 16, 32\} \\
Stalk dim $d$           & \{2, 3, 4\} \\
Layers                  & \{2, 3, 4, 5, 6\} \\
Learning rate           & 0.02 \\
Activation              & ELU \\
Weight decay (regular)  & Log-uniform $[10^{-9.2}, 10^{-6.8}]$ \\
Weight decay (sheaf)    & Log-uniform $[10^{-11.0}, 10^{-6.8}]$ \\
Input dropout           & Uniform $[0.0, 0.9]$ \\
Layer dropout           & Uniform $[0.0, 0.9]$ \\
Add LP / Add HP         & \{0, 1\} (separately for each) \\
KL term                 & Enabled \\
Ensemble size           & 3 \\
Orthogonalization       & Householder \\
Patience (early stopping) & 200 \\
Max training epochs     & 1000 \\
Optimizer               & Adam \\
\bottomrule
\end{tabular}
\caption{Hyperparameter ranges for BSNN models.}
\label{tab:config}
\end{table}

To perform a hyperparameter sensitivity analysis, we used the hyperparameters listed in Table \ref{tab:hyp} with node splits of 32\% for training, 20\% for validation, and 48\% for testing.
\begin{table*}[htbp]
    \centering
    \begin{tabular}{l | l} 
     \hline 
     \textbf{Hyperparameter} & \textbf{Values} \\
     \hline
     \rule{0pt}{1em}Hidden channels & \{8, 32\} \\ 
     Stalk dim $d$ & \{2, 3, 4, 5\} \\
     Layers & \{2, 3, 4, 5\} \\
     Dropout &  \{0.0, 0.3, 0.6\}  \\
     Learning rate & 0.01  \\
     Activation & ELU  \\
     Weight decay & 5e-4  \\
     Patience & 200  \\
     Max training epochs &  500  \\
     Optimiser & Adam \\ 
     Num. models in ensemble & 3 (BSNN), N/A (NSD) \\ [0.5ex] 
     \hline
    \end{tabular}
        \caption{Hyperparameter selections for sensitivity analysis using 32\%/20\%/48\% node splits for training, validation, and testing.}
    \label{tab:hyp}
\end{table*}

\end{document}